\documentclass[conference]{IEEEtran}  

\IEEEoverridecommandlockouts                              

\usepackage{cite}
\usepackage[T1]{fontenc}
\usepackage[utf8]{inputenc}
\usepackage{authblk}
\usepackage{verbatim}
\usepackage{bm}
\usepackage{amsmath,amssymb,amsfonts}
\usepackage{algorithmic}
\usepackage{subfigure}
\usepackage{graphicx}
\usepackage{textcomp}
\usepackage{xcolor}
\usepackage{multirow}
\usepackage{amsmath} 
\usepackage{amssymb}
\usepackage{wrapfig}
\usepackage{caption}
\usepackage[ruled,linesnumbered]{algorithm2e}
\def\BibTeX{{\rm B\kern-.05em{\sc i\kern-.025em b}\kern-.08em
    T\kern-.1667em\lower.7ex\hbox{E}\kern-.125emX}}
\title{\LARGE \bf
Learning-Based Motion Planning with Mixture Density Networks
}

\author{Yinghan Wang, Xiaoming Duan, and Jianping He
\thanks{The authors are with the Department of Automation, Key Laboratory of System Control and Information Processing, Shanghai Jiao Tong University, Shanghai, China, 200240.
Email:\small{\{wyhboos, xduan,  jphe\}@sjtu.edu.cn}}}

\begin{document}
\bstctlcite{IEEEexample:BSTcontrol}

\maketitle

\begin{abstract}
The trade-off between computation time and path optimality is a key consideration in motion planning algorithms. While classical sampling based algorithms fall short of computational efficiency in high dimensional planning, learning based methods have shown great potential in achieving time efficient and optimal motion planning. The SOTA learning based motion planning algorithms utilize paths generated by sampling based methods as expert supervision data and train networks via regression techniques. 
However, these methods often overlook the important multimodal property of the optimal paths in the training set, making them incapable of finding good paths in some scenarios.  In this paper, we propose a Multimodal Neuron Planner (MNP) based on the mixture density networks that explicitly takes into account the multimodality of the training data and simultaneously achieves time efficiency and path optimality. For environments represented by a point cloud, MNP first efficiently compresses the point cloud into a latent vector by encoding networks that are suitable for processing point clouds. We then design multimodal planning networks which enables MNP to learn and predict multiple optimal solutions. Simulation results show that our method outperforms SOTA learning based method MPNet and advanced sampling based methods IRRT* and BIT*.
\end{abstract}

\section{Introduction}
\label{sec:intro}
\par Motion planning is one of the core research topics in robotics \cite{2015_ICIA_Peng_obstacle_avoidance}. Motion planning algorithms aim to find a collision free path in a specific environment given the initial and goal configuration of the robot. The key desirable properties of a motion planning algorithm include: (1) the algorithm is guaranteed to generate a collision free path if it exists (completeness), and the path found by the algorithm should have the lowest path cost (optimality); (2) the algorithm is expected to find a path in real time (time efficiency); (3) the algorithm should be able to generate plans in high dimensional configuration spaces (generalization to high dimensions).

\par Numerous motion planning algorithms have been developed in the past decades, such as the artificial potential fields method \cite{1986_ARV_Khatib_APF}, the vector field histogram method \cite{1991_Borenstein_TRA_VFH} and their improved versions \cite{2015_ICRA_Chiang_APF_m} \cite{2000_ICRA_Ulrich_VFH*}. These algorithms are simple and useful in simple environments but not suitable for high dimensional planning. Sampling based methods such as RRT \cite{1998_Ames_LaValle_RRT} try to find a collision free path through repetitively sampling in the configuration space. RRT* \cite{2011_IJRR_Karaman_RRT*}, IRRT* \cite{2014_IROS_Gammell_IRRT*} and BIT* \cite{2015_ICRA_Gammell_BIT*} are further developed to improve the optimality of the generated paths. While sampling based methods are able to find a collision free and near-optimal path and can be used to plan in high dimensions, there is a trade-off between the computing time and optimality. It often takes a lot of time for the sampling based methods to find an optimal path,  which makes them not suitable for real time planning.

\par With the advancement of machine learning techniques, learning based motion planning algorithms have been developed and have shown promising performance. To improve the efficiency for sampling based methods, the authors in \cite{2018_ICRA_Ichter_Learning_sampling_distributions} and \cite{2018_IROS_Zhang_Learning_implicit_sampling} develop algorithms that learn and predict the sample distributions to guide the sampling process. The work \cite{2019_IROS_Bency_Neural_path_planning}, \cite{2020_TRO_Qureshi_Motion_planning_networks} and \cite{2019_ICRA_Qureshi_Motion_planning_networks} utilize deep networks that learn from expert experiences and predict motions iteratively to improve the efficiency of planning. However, there are still problems with existing methods. The ability to process environment information properly is the key for generalization and the point clouds are suitable to represent obstacles in environments, while \cite{2019_IROS_Bency_Neural_path_planning} does not take the environment information as the input and \cite{2020_TRO_Qureshi_Motion_planning_networks} \cite{2019_ICRA_Qureshi_Motion_planning_networks} process point clouds in an inefficient way. On the other hand, it is common that there are multiple solutions for motion planning problems, but \cite{2020_TRO_Qureshi_Motion_planning_networks} and \cite{2019_ICRA_Qureshi_Motion_planning_networks} lack the ability to learn and generate multiple solutions which cause serious problems when learning from expert experiences.

\par In this paper, we propose a learning based motion planning algorithm called Multimodal Neural Planner (MNP) which takes in point clouds, robot's initial and goal configuration and plans a near-optimal path efficiently by iteratively generating the next configuration. The planner can learn and generate multiple solutions, and it consists of two networks. The environment encoding networks utilize the maxpooling mechanism similar to \cite{2017_CVPR_Qi_PointNet} to compress the input point clouds into a latent vector efficiently, and the planning networks take in the latent vector and the robot's current and goal configuration and predict the next configuration. Inspired by \cite{1994_AU_Bishop_MDN}, we describe the next configuration by a Gaussian mixture model whose parameters are determined by the output of planning networks. The Gaussian mixture model enables the planning networks to learn and predict multiple solutions. Based on the above two networks, we further utilize the bi-directional planning and the replanning mechanism to form a complete, near-optimal and efficient planner. Simulation results show that our method is able to solve motion planning problems in high dimensions and outperforms the SOTA learning based planner MPNet~\cite{2020_TRO_Qureshi_Motion_planning_networks} and advanced sampling based planners such as IRRT* and BIT*.


\section{Related Works}
\label{sec:related}
\par 

Imitation learning is one of the popular ways to solve motion planning problems through learning
from expert’ experiences. Many of the existing imitation learning methods adopt the end-to-end training framework to generate control commands from the input sensor information. In \cite{2018_IROS_Liu_Map-based_deep_imitation} \cite{2017_IVS_Kuefler_imitate_GAN} \cite{2017_ICAR_Pfeiffer_From_perception_to_decision} \cite{2016_IROS_Tai_model-less_obstacle_avoidance}, the mobile robot learns from the expert's demonstrations with supervised learning and imitates to generate the control command that  navigates the robot to the goal while avoiding obstacles. The expert's demonstrations are either generated by humans \cite{2017_IVS_Kuefler_imitate_GAN} \cite{2016_IROS_Tai_model-less_obstacle_avoidance} or by classical controllers \cite{2018_IROS_Liu_Map-based_deep_imitation} \cite{2017_ICAR_Pfeiffer_From_perception_to_decision}. Several works also study how to improve the performance of the generated control command. In \cite{2018_ICRA_Codevilla_Conditional_Imitation_Learning}, the authors design a command-conditional controller that generates the steering command at road intersections. Active learning has been utilized in~\cite{2019_Xiong_Learning_Safety-Aware} to improve the ability to avoid obstacles which enhances safety. In \cite{2017_CoRL_Gao_Intention-net}, the low level controller is instructed by the high level planned route through the intention-net.  In this work, we focus on planning a global path instead of designing learning based controllers.

\par Since sampling based methods are complete and optimal, they are widely used in the motion planning. To improve the time efficiency of sampling based methods, methods have been developed to guide the sampling process. For example, in \cite{2020_TASE_Wang_Neural_RRT*} \cite{2020_ICRA_Ichter_Learned_critical_probabilistic_roadmaps} a probability map is generated for sampling and in \cite{2021_ARXIV_Johnson_Motion_planning_transformers} transformer is utilized to mark out the sampling area.  However, \cite{2020_TASE_Wang_Neural_RRT*} \cite{2021_ARXIV_Johnson_Motion_planning_transformers} are only suitable for environments in low dimensions. In \cite{2018_ICRA_Ichter_Learning_sampling_distributions} \cite{2018_IROS_Zhang_Learning_implicit_sampling}, the sample distribution has been predicted. In \cite{2018_IROS_Qureshi_Deeply_informed_neural_sampling} the sample location is directly predicted, but it is unable to learn and predict multiple solutions. In \cite{2019_ARXIV_Chen_NEXT} deep networks are used to help select the parent node and expand node of the search tree, it takes environment map as input while our method uses point clouds which is easier to obtain by sensors. By contrast, we propose a method that directly generates the path iteratively without using a search tree, and it can generate multiple solutions and be applied in high dimensional environments.

\par The deep networks are also utilized to directly generate paths for motion planning problems. In \cite{2021_CRV_Toma_Waypoint_planning_networks}, the Waypoint Planning Networks predicts waypoints with deep networks and connects them by classic methods such as A*. In \cite{2020_ICIP_Watt_Pathnet}, the Pathnet directly generates the path by Generative Adversarial Networks~\cite{2020_ACM_Goodfellow_GAN}. However, the above two methods are suitable only for 2D planning. In \cite{2019_IROS_Bency_Neural_path_planning} the path is generated by iteratively predicting the next configuration with LSTM~\cite{1997_NC_Hochreiter_LSTM}, but it does not take environment information as input which makes it unable to generalize to new environments.

\par The Motion Planning Networks (MPNet) \cite{2019_ICRA_Qureshi_Motion_planning_networks} \cite{2020_TRO_Qureshi_Motion_planning_networks} take the point clouds as input and iteratively predict next configuration to form a path, which is similar to our method. However, there are two main problems with MPNet. First, MPNet encodes point clouds with fully connected networks which is not suitable for processing disordered cloud points. Second, MPNet learns from expert's data using regression techniques, and it cannot learn from and predict multiple solutions. Our method overcomes these critical issues.
PG-RRT~\cite{2022_TASE_Lyu_points-guided_sampling_network} is another sampling based method with deep networks that uses mixture density networks to generate sampling distributions similar to ours. However, PG-RRT aims to solve the discontinuity in motion planning, and our method enables networks to learn and predict multiple solutions. Also, PR-RRT is a sampling based method while ours generates the path without building a search tree.

In this paper, we mainly reveal an important yet overlooked issue with existing solutions to learning-based motion planning problems. That is, when learning from the expert data generated by sampling based methods, existing methods do not consider the fact that multiple different solutions with similar optimality could coexist in the data set. Due to this, they cannot mimic the behavior of the expert sampling based methods. To address this issue, we propose a mixture density network based multimodal planning network, which learns from the training data with multimodal property and can produce multiple solutions.


\section{Problem Formulation and Motivation}
\label{sec:definition}
\subsection{Problem formulation}
\par Let the robot configuration space be $C \subset \mathbb{R}^{d}$ where $d\in\mathbb{N}$ is the dimension of the configuration space. The configuration space $C$ is comprised of the collision space $C_{\mathrm{col}}$ and the free space $C_{\mathrm{free}}$ where $C_{\mathrm{free}}=C\setminus C_{\mathrm{col}}$. The robot's physical environment space is called the workspace and is denoted by $W\subset \mathbb{R}^{m}$ where $m$ is the dimension of the workspace. Similarly, the workspace is comprised of the collision space $W_{\mathrm{obs}}$ (regions occupied by obstacles) and the free space $W_{\mathrm{free}}$ where $W_{\mathrm{free}}=W\setminus W_{\mathrm{obs}}$.
In the workspace $W$, the robot also occupies a physical region that depends on its configuration $c\in C$. Collision occurs when $W_{\mathrm{obs}}$ and the region occupied by robot overlaps. In the case of collision, we have that $c\in C_{\mathrm{col}}$, otherwise $c\in C_{\mathrm{free}}$. Let $\Phi:C\to \{\mathrm{False}, \mathrm{True}\}$ be the collision check function that takes in the robot's configuration and outputs whether there is collision.
\par Let the robot's initial configuration be $c_{\mathrm{init}}\in C_{\mathrm{free}}$ and the robot's goal configuration space be $C_{\mathrm{goal}}\subset C_{\mathrm{free}}$, then the motion planning problem can be described as follows. Given the initial configuration and the goal configuration space, find a path represented by an ordered list $\tau =\left \{ c_0,\dots,c_i,\dots,c_l \right \}$ such that $c_0 = c_{\mathrm{init}}$, $c_l \in C_{\mathrm{goal}}$ and $c_i\in C_{\mathrm{free}}$ for $i = 0,\dots,l$. Moreover, the line segment between $c_i$ and $c_{i+1}$ should lie entirely in the free configuration space $C_{\mathrm{free}}$. In practice, the interpolation method is usually used to check if the line segment is collision free: it first computes the intermediate configurations between $c_i$ and $c_{i+1}$ and then utilizes the collision check function $\Phi$ to determine if all those intermediate configurations are collision free. A path that satisfies the conditions above is a feasible solution to the motion planning problem. Sometimes finding a feasible path is not enough, and we can evaluate the path quality and obtain a best one based on a cost function. Therefore, a good planning algorithm is expected to find a path that has the minimum cost value as measured by some cost function.

\begin{figure}[t]
\centering
\includegraphics[page=21,width=2.5in]{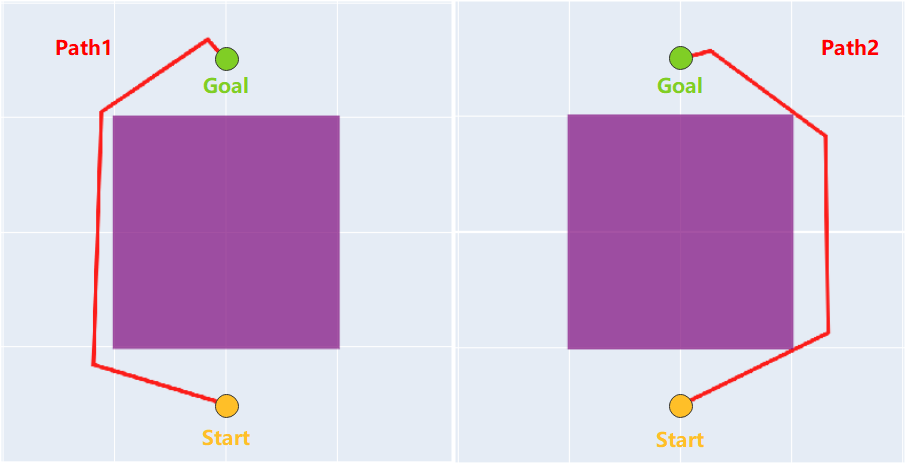}
\caption{Two different but near-optimal solutions for a simple motion planning problem. The bottom is the initial configuration and the top is the goal point.}
\label{fig:multi_solution}
\end{figure}

\subsection{Multimodal property of motion planning solutions}
\label{sec:multimodal property}


\par 
The solution path generated by  motion planning algorithms is generally not unique, i.e., there may exist different solution paths that have the same lowest cost value and all of them are in the solution set of the problem.
We call such property of the motion planning problems the multimodal property. Fig.~\ref{fig:multi_solution} depicts a scenario where there are two completely different but near-optimal solutions generated by RRT* for a simple 2D planning problem.

\par Since deep imitation learning based motion planning algorithms learn from the expert's experience, their performance is profoundly affected by their ability to fit the expert data. In the above situation, the deep imitation learning method should learn all the choices that can possibly be made by the expert (e.g., RRT*). 
However, existing methods in \cite{2018_IROS_Qureshi_Deeply_informed_neural_sampling, 2019_IROS_Bency_Neural_path_planning, 2019_ICRA_Qureshi_Motion_planning_networks, 2020_TRO_Qureshi_Motion_planning_networks} which use a simple mean squared error (MSE) loss function to train the network are not capable of generating multimodal solutions. As mentioned in Section \uppercase\expandafter{\romannumeral2}, MPNet obtains a path by iteratively predicting the next configuration $c_{\mathrm{next}}$ based on the current configuration $c_{\mathrm{current}}$, goal configuration $c_{\mathrm{goal}}$ and the environment information $Z$. In MPNet, the training data set generated by a sampling based method such as RRT* consists of tuples $(c_{\mathrm{current}}, c_{\mathrm{goal}}, Z, c_{\mathrm{next}})$ where $c_{\mathrm{current}}$, $c_{\mathrm{goal}}$, $Z$ are inputs and $c_{\mathrm{next}}$ is the output. During the training process, the following MSE loss function is utilized
\begin{equation}
\min_{\theta} \sum_{i=1}^{N} \left || \hat{c_i}(\theta)-  c_i\right || _2^{2},
\label{eq:MPN_Pnetloss}
\end{equation}
where $\theta$ is the parameters of the planning networks, $N$ is the number of training data, $\hat{c_i}(\theta)$ and $c_i$ are the predicted and labeled configurations, respectively. While the MSE loss is simple and intuitive, it restricts the planning networks to generate only one solution for the motion planning problem. Moreover, when there are more than one solution in the training dataset, networks with the MSE loss fail to learn any of the solutions because an "average solution" leads to a minimum cost. 
For the simple 2D path planning problem in Fig.~\ref{fig:multi_solution} where the initial position is at the bottom and the goal position is at the top, two different optimal paths exist in the training data. With the MSE loss, the planning networks can only learn the average of the paths and outputs the decision of going up north at the initial position. This is unacceptable because it leads to collision with the obstacle. To alleviate this issue, MPNet utilizes dropout, which is proven to be a Bayesian approximation in \cite{2016_ICML_Gal_dropout_as_bayesian}, to promote randomness. However, as shown in Fig.~\ref{fig:multimodal comparison} in our later experiment section, MPNet fails to predict the multimodal solution when the size of the training data becomes larger, where a $0.5$ dropout probability is applied to each layer of the neural network~\cite{2020_TRO_Qureshi_Motion_planning_networks}. Overall, MPNet with the MSE loss is not able to correctly learn and generate multiple solutions and can result in error when learning from multiple solutions.

\par In practice, it is quite common to get multiple solutions for one motion planning problem when generating training data using sampling based method. 
The multimodal problem not only exists in situations shown in Fig.~\ref{fig:multi_solution} where there are two choices at the starting position, but it can also show up whenever multiple equally valid choices are available at a given location. 
To make planning networks smarter and more efficient, we need to develop a method that overcomes the above problems and allows the planning networks to learn and predict multiple solutions. The proposed multimodal planning networks in Section~\ref{sec:method Pnet} have the ability to learn multiple solutions by utilizing Mixture Density Networks.
\section{Method}
\label{sec:method}

\begin{figure*}[t]
\centering
\includegraphics[scale=0.4]{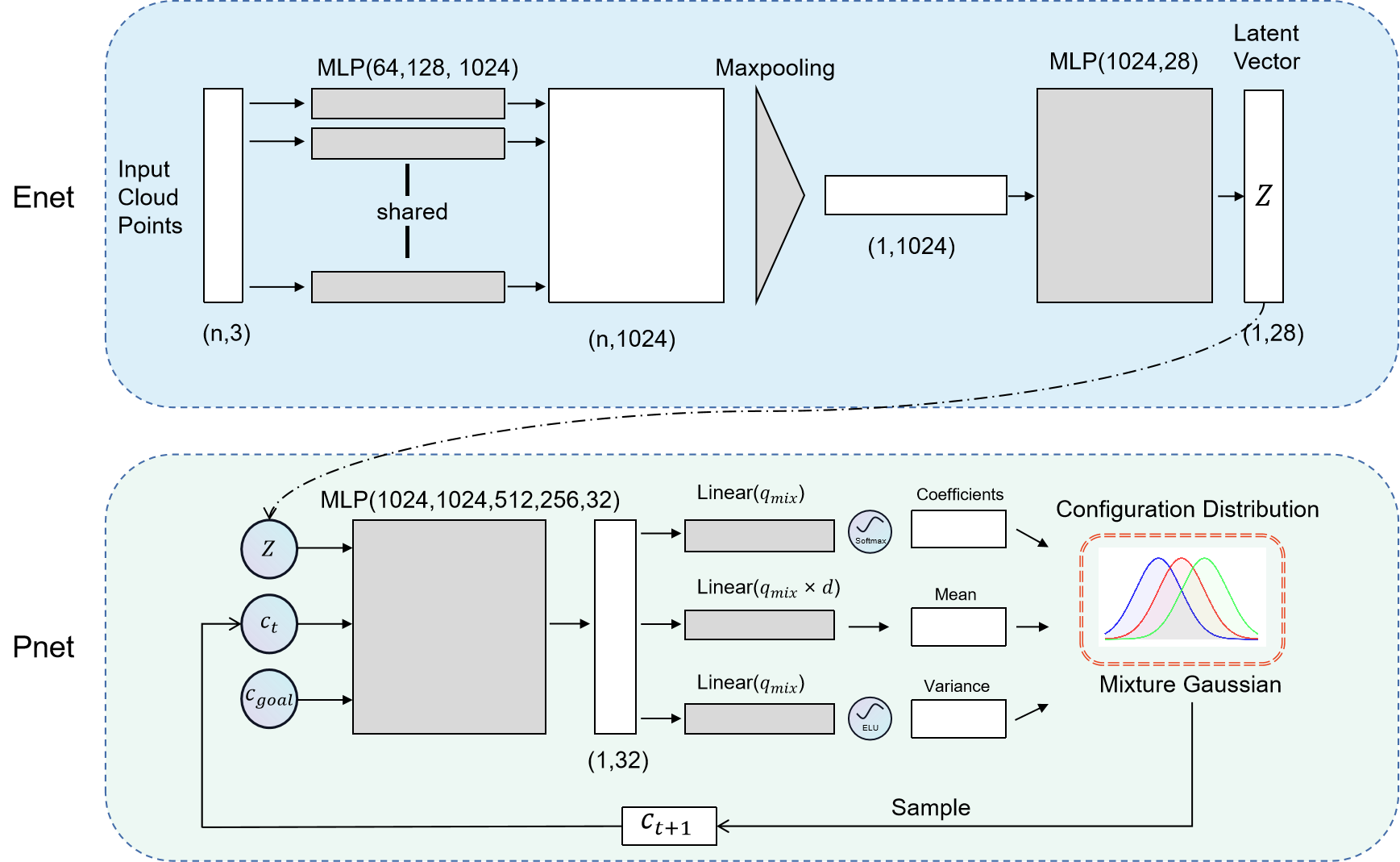}
\caption{The overview of the structure of the proposed method. Enet takes $n$ cloud points as input where 
 the dimension of points is $3$ for a 3D environment and $2$ for a 2D environment.  Each point is processed by the same MLP. Then, a maxpooling operation is performed along the dimension of the number of cloud points $n$. Finally, another MLP is applied to obtain the latent vector $Z$. Pnet takes in the latent vector $Z$, the current configuration $c_t$ and the goal configuration $c_\mathrm{goal}$, through MLP and specific activation functions, and it generates the coefficients, mean and variance of the mixture Gaussian distribution with which we can obtain the next configuration $c_{t+1}$ by sampling. Finally, we replace $c_{t}$ with $c_{t+1}$ and run Pnet iteratively to obtain a complete path.} 
\label{fig:structure}
\end{figure*}

\par The structure of the proposed method is shown in Fig.~\ref{fig:structure}. The proposed multimodal Motion Planning Networks consists of two parts, environment encoding networks (Enet) and the planning networks (Pnet). The Enet aims to encode the environment information which is represented by point clouds into a latent vector to help the following Pnet better understand the environment. We design our Enet by referring to PointNet \cite{2017_CVPR_Qi_PointNet} which is well suited for processing point clouds. The Pnet takes the latent vector obtained from the Enet as input and iteratively generates waypoints to form a path. As the expert data are multimodal, we utilize the mixed density networks \cite{1994_AU_Bishop_MDN} to learn from the expert and generate waypoints.
\subsection{Environment encoding networks}
\label{sec:method Enet}

\par The use of point clouds to represent environment information is practical and common \cite{2020_TRO_Qureshi_Motion_planning_networks}. The point clouds characterize the geometric features of the environment, and
extracting and understanding those features is very important for the downstream motion planning task. In this work, we develop an environment encoding network to process the point clouds and convert them into a latent vector
\begin{equation}
Z = \mathrm{Enet}(X_{\mathrm{cloud}};\theta_{e}),
\label{eq:Enet}
\end{equation}
where $X_{\mathrm{cloud}}=\left \{  x_1,\dots,x_n\right \}$ is the set of cloud points with $x_i \in W$, and $\theta_e$ is the parameters of the Enet.

\par

To learn the environment representation, MPNet utilizes the multi-layer perception (MLP) to process the point clouds.
However, since cloud points do not have a fixed order, MLP is not an efficient method to process them. 
To overcome this issue and similar to~\cite{strudel2021learning}, we use the PointNet~\cite{2017_CVPR_Qi_PointNet} to obtain global features, where a shared MLP and a maxplooling layer are utilized.



\par The architecture of the Enet is shown in Fig.~\ref{fig:structure}. For the training of the Enet, unlike \cite{2022_TASE_Lyu_points-guided_sampling_network, 2018_IROS_Qureshi_Deeply_informed_neural_sampling, 2019_ICRA_Qureshi_Motion_planning_networks, 2020_TRO_Qureshi_Motion_planning_networks}, we do not use the encoder-decoder approach with unsupervised learning. Instead, we train the Enet and the Pnet together. The reason for this is that the Enet should be well informed of the downstream motion planning task in order for it to retain critical environmental features relevant to the specific task, and when trained together with the Pnet, it receives error feedback from the output of the Pnet based on which the parameters can be adjusted. On the other hand, our ultimate goal is not to solely minimize the reconstruction error of the environment as in the encoder-decoder approach. 

\subsection{Multimodal planning networks}
\label{sec:method Pnet}
\par After obtaining the latent vector $Z$ through the Enet, the multimodal Pnet is designed to generate a full path iteratively. The architecture of the Pnet is also shown in Fig.~\ref{fig:structure}. The Pnet takes in $Z$, the current configuration $c_t$ and the goal configuration $c_{\mathrm{goal}}$, and it predicts the next configuration $c_{t+1}$. Then,  $c_t$ is replaced  with $c_{t+1}$ and a full path is generated iteratively. As mentioned in Section~\ref{sec:multimodal property}, there are multiple different solutions to a motion planning problem, and the Pnet must be able to learn and predict these solutions. To achieve this, we utilize a probability distribution to describe the predicted next configuration. Specifically, we adopt the Mixture Density Networks (MDN) which utilizes deep networks to represent the parameters of the Gaussian Mixture Model (GMM) as the backbone of our Pnet. In our Pnet, MDN predicts the distribution of the next configuration with GMM that is comprised of multiple Gaussian distributions. The GMM is determined by the following parameters: coefficients $\alpha_i$ of each Gaussian distribution, mean $\mu_i$ of each Gaussian distribution and variance $\sigma_i$ of each Gaussian distribution. Note that we assume that each component of the output vector is independent and its variance is described by a common $\sigma_i$ for each Gaussian. Under such an assumption, the GMM can still approximate any distribution \cite{1994_AU_Bishop_MDN} given enough Gaussians. Since the parameters $\alpha$ and $\sigma$ must satisfy certain constraints, different activation functions are used. We apply the Softmax activation to obtain $\alpha$ so that each $\alpha_i$ is nonnegative and the sum of $\alpha_i$'s is one. For $\sigma$, we use the ELU activation instead of the EXP activation for better numerical statbility \cite{2017_UP_Brando_MDN_for_distribution_and_uncertainty_estimation}. The parameters $\mu$ are obtained directly from the output of networks. Through the Pnet, we can obtain the distribution of the robot's next configuration as 
\begin{equation}
p(c|\alpha, \mu, \sigma) = \sum_{i=1}^{q_{\mathrm{mix}}} \alpha _i\frac{1}{(2\pi)^{d/2}\sigma_i^d}e^{-\frac{||c-\mu_i||^2 }{2\sigma_i^2 } }, \label{eq:GMM_prob}
\end{equation}
where $c$ is the robot configuration in dimension $d$, and $q_{\mathrm{mix}}$ is the number of Gaussian distributions. The parameters $\alpha$, $\mu$ and $\sigma$ are the output of the Pnet
\begin{equation}
(\alpha, \mu, \sigma) = \mathrm{Pnet}(c_{t}, c_{\mathrm{goal}}, Z;\theta_p),
\label{eq:Pnet_output}
\end{equation}
where $\theta_p$ is its parameters. Finally, we sample from \eqref{eq:GMM_prob} to obtain the next configuration
\begin{equation}
\hat{c}_{t+1} = \mathrm{Sample}(p(c|\alpha, \mu, \sigma)).
\label{eq:sample}
\end{equation}
As discussed in Section~\ref{sec:method Enet}, we train both the Enet and the Pnet together in a supervised end-to-end manner. Since the Pnet outputs a probability distribution, the negative logarithmic likelihood (NLL) is adopted as the loss function, and the optimization problem can be described by
\begin{equation}
\min_{\theta_e,\theta_p} - \sum_{i=1}^{N}  \ln p(c_{t+1}),
\label{eq:loss}
\end{equation}
where $N$ is the number of training data points and $p(c_{t+1})$ is obtained from \eqref{eq:GMM_prob} and \eqref{eq:Pnet_output}. The training data is organized into the form of $(c_{t}, c_{\mathrm{goal}}, X_{\mathrm{cloud}}, c_{t+1})$ with $(c_{t}, c_{\mathrm{goal}}, X_{\mathrm{cloud}})$ as input and $c_{t+1}$ as output, and the training data is acquired from sampling based methods such as RRT* with sufficient running time. More training details will be discussed in Section~\ref{sec:results}. The proposed motion planning method Multimodal Neuron Planner (MPN) is composed of the Enet and the Pnet. It uses a similar algorithmic framework to MPNet with some improvements, and we include the algorithm details in the appendix.

\section{Results}
\label{sec:results}
\par We present the results of simulation studies in this section. First, we show the multimodal properties of the proposed MNP by visualizing the trained GMM in Section~\ref{sec:multimodal comparison}. Then, we comprehensively compare the MNP with the SOTA MPNet and advanced sampling based methods such as BIT* and Informed-RRT* (IRRT*) using multiple different robot models in different environments in Section~\ref{sec:comparison multiple}. Finally, we compare above algorithms for the motion planning task of a 7-DOF Panda Arm in Section~\ref{sec:comparison panda arm}.

\subsection{Comparison of algorithms' multimodal properties}
\label{sec:multimodal comparison}
\begin{figure}[htbp]
\centering
\subfigure[MNP: case 1.]{
\includegraphics[width=1.15in]{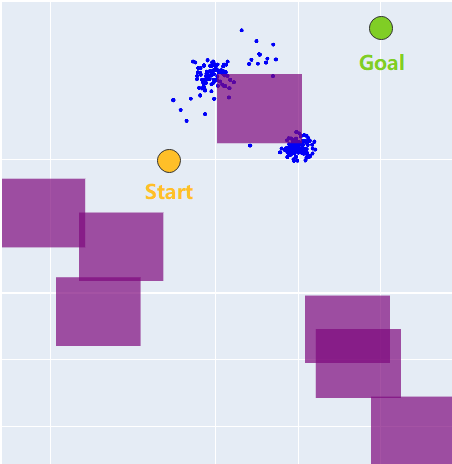}
}
\hspace{-0.4cm}
\subfigure[MNP: case 2.]{
\centering
\includegraphics[width=1.15in]{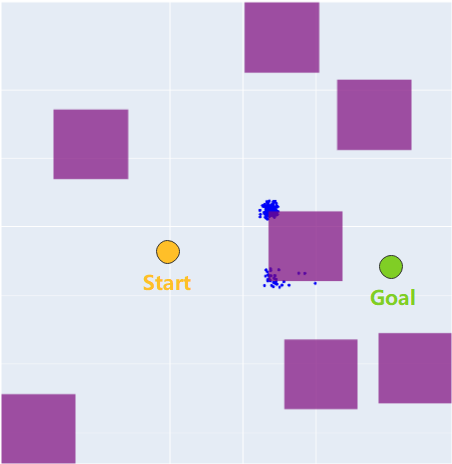}
}
\hspace{-0.4cm}
\subfigure[MPNet: case 1.]{
\centering
\includegraphics[width=1.15in]{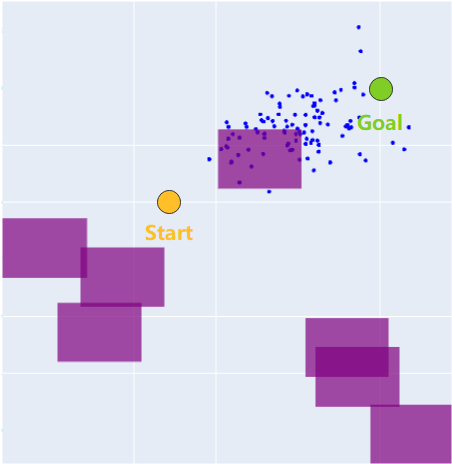}
}
\hspace{-0.4cm}
\subfigure[MPNet: case 2.]{
\centering
\includegraphics[width=1.15in]{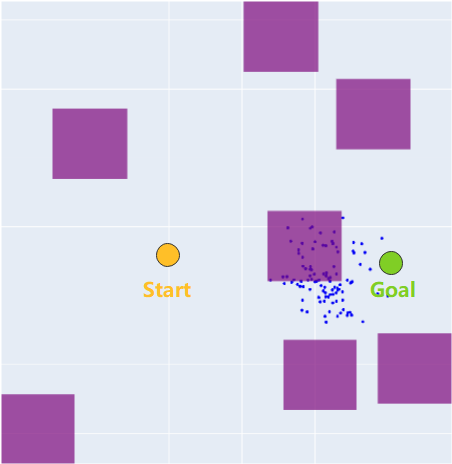}
}
\caption{Distributions of the predicted next configuration by MNP and MPNet which are represented by blue scatters, and the obstacles are marked in purple. For MNP, the distribution is obtained by repetitive sampling, and for MPNet, it is obtained by multiple forward propagations with the train mode on in Pytorch to introduce the stochastic property of the dropout layer.}
\label{fig:multimodal comparison}
\end{figure}
\par To evaluate the ability of our multimodal Pnet to learn and generate multiple solutions, we repeatedly generate the next configuration 
for a point robot model in a simple 2D environment and plot the distribution. As shown in Fig.~\ref{fig:multimodal comparison}, MNP generates multiple reasonable
next configurations, while MPNet generates relatively monotonous configurations that lie around the straight line to the goal even when the dropout layer is applied. We also compare the collision rate of the generated configurations which profoundly influences the algorithm's efficiency. Our multimodal Pnet has a lower collision rate at $20\%$ while MPNet reaches $42\%$. The above shows that MNP can learn and predict multiple solutions and performs better than MPNet.


\subsection{Comparisons using multiple different robot models}
\label{sec:comparison multiple}

\begin{figure*}[htbp]
\centering
\subfigure[2D point.]{
\includegraphics[width=1.2in]{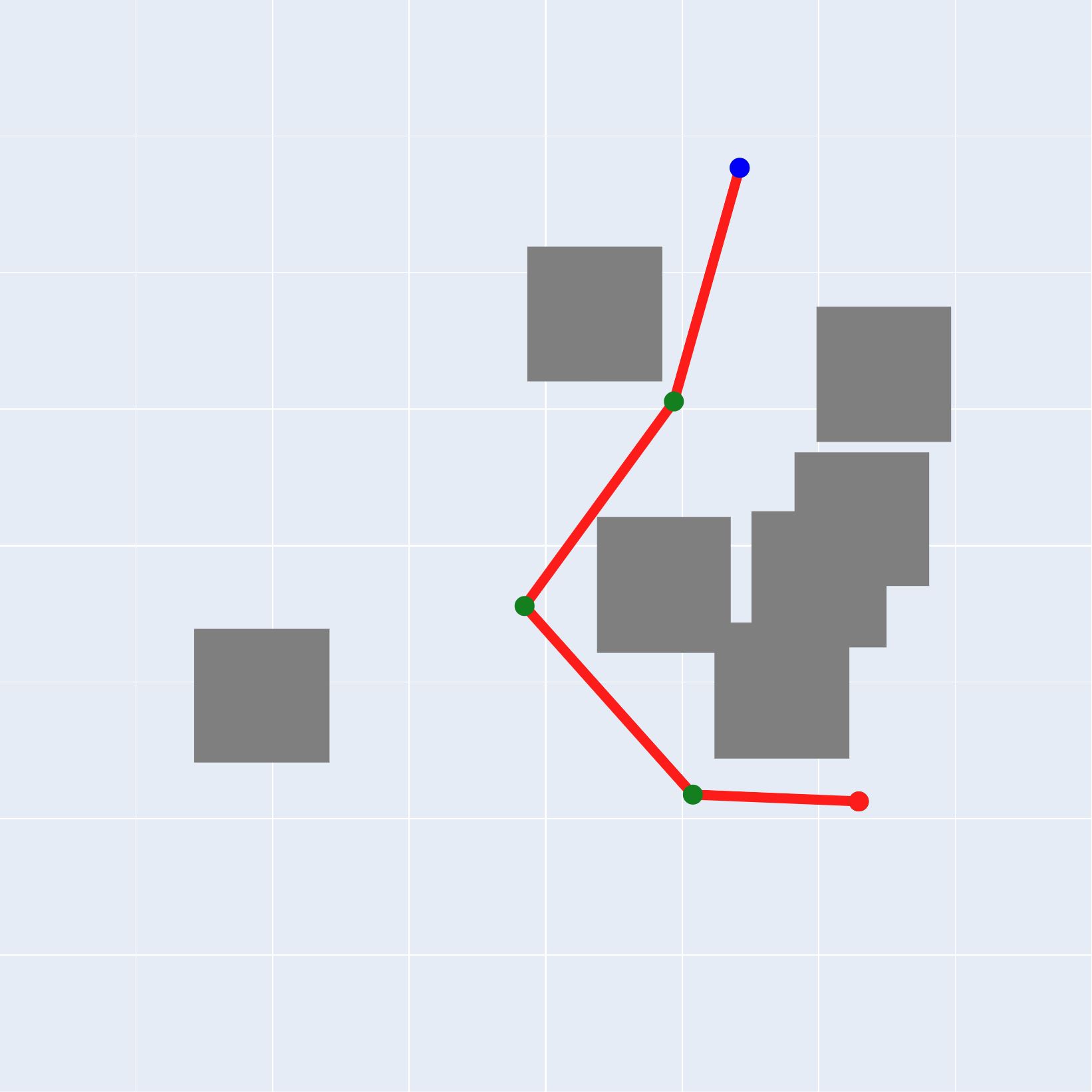}
}
\subfigure[2D rigid.]{
\includegraphics[width=1.2in]{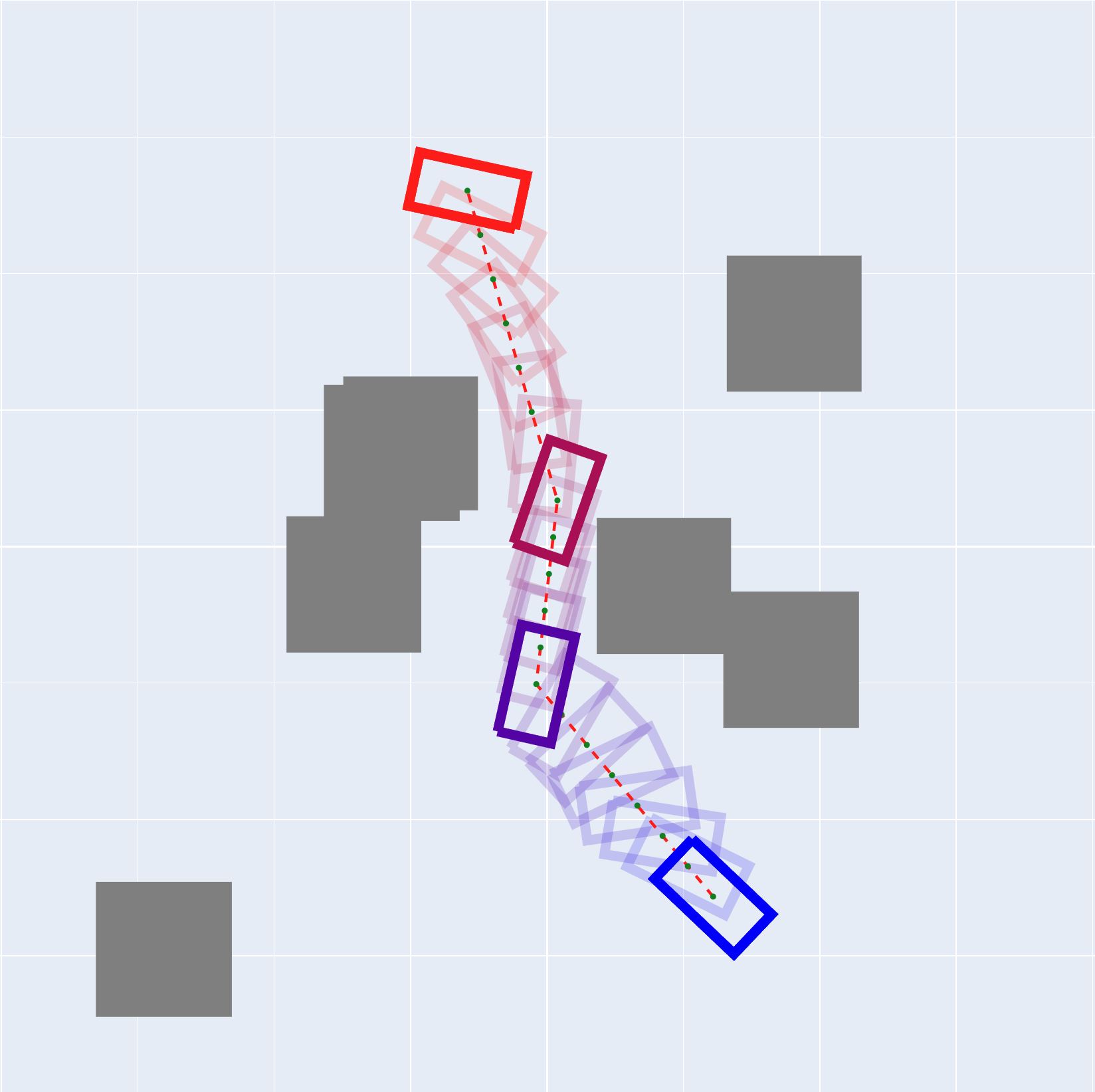}
}
\subfigure[2D 2-link.]{
\centering
\includegraphics[width=1.2in]{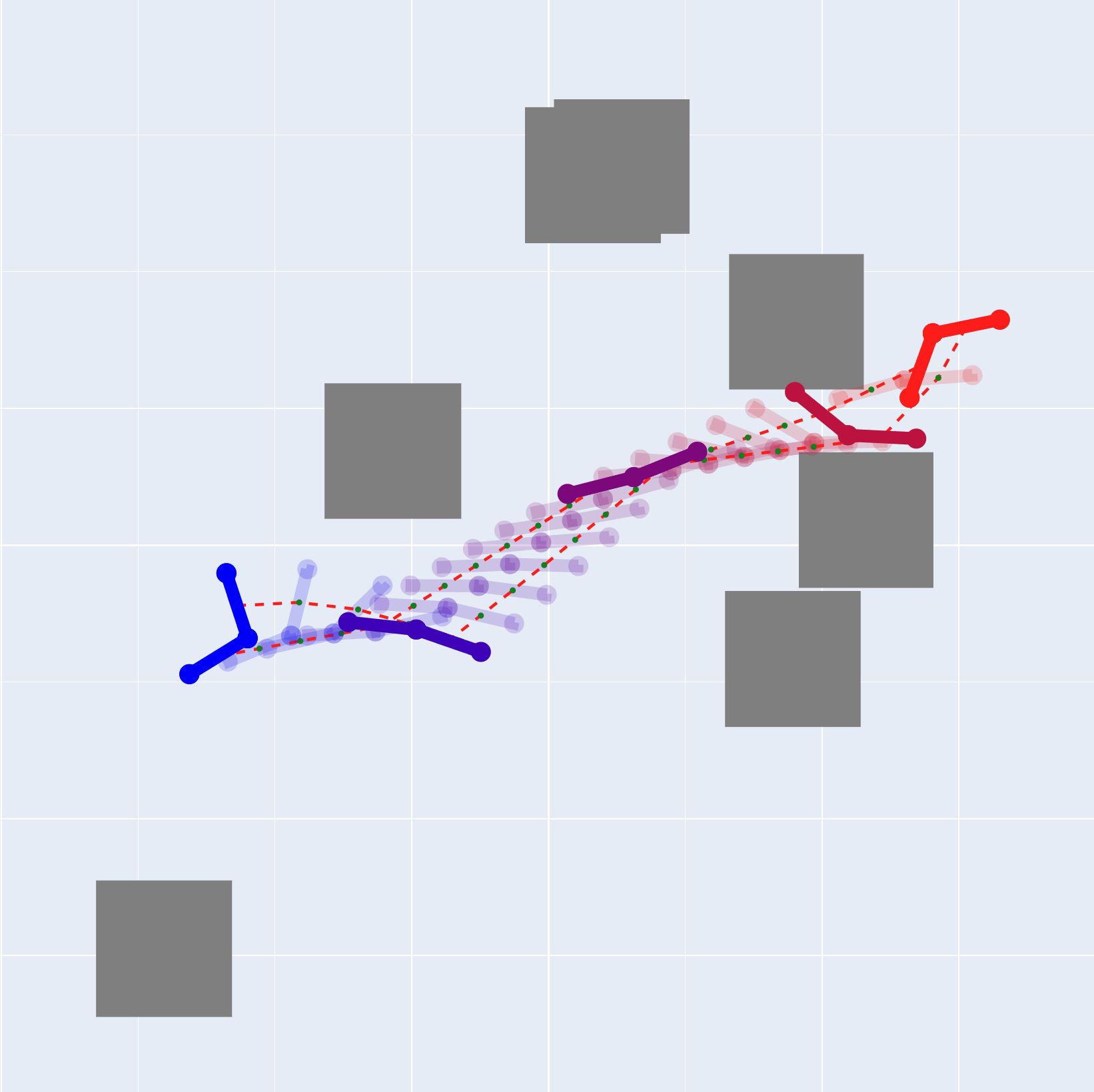}
}
\subfigure[2D 3-link.]{
\centering
\includegraphics[width=1.2in]{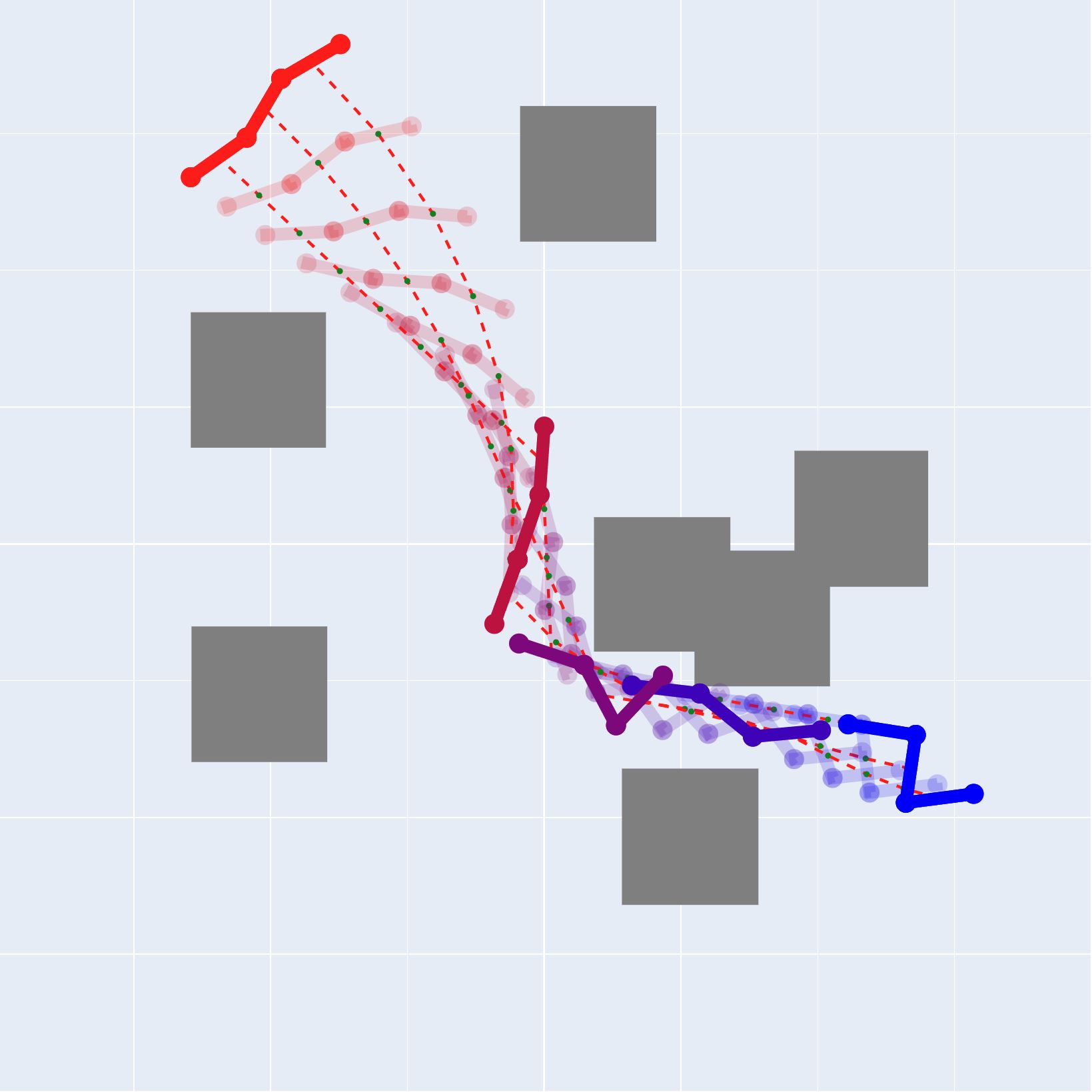}
}
\subfigure[3D point.]{
\includegraphics[width=1.2in]{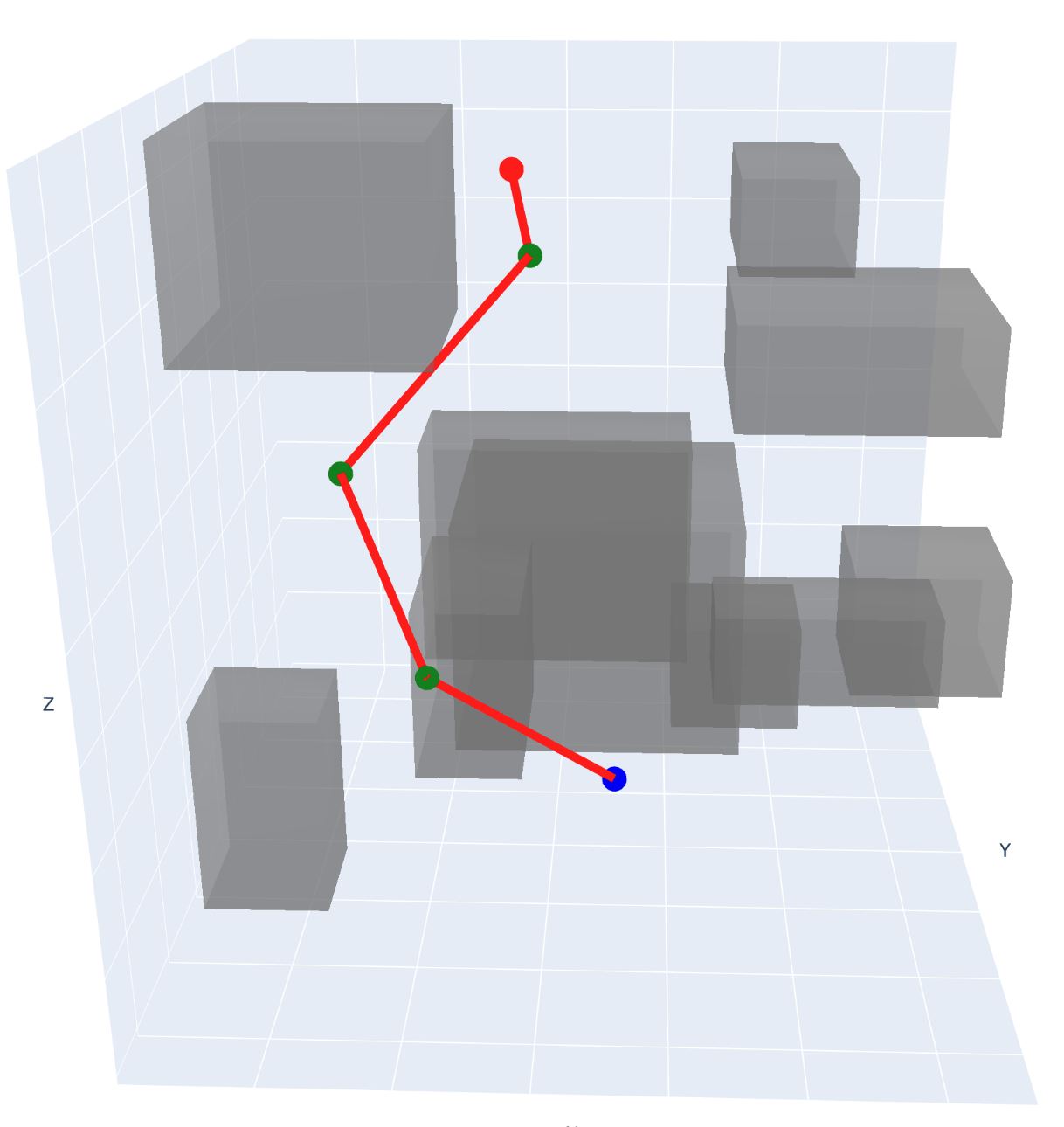}
}
\caption{Paths planned by MNP with different environment setup. Initial and goal configurations are marked as blue and red, respectively.}
\label{fig:multi strcuture}
\end{figure*}


\begin{small}
    \begin{table*}[!ht]
\center
\begin{tabular}{|c|c|c|c|c|c|c|c|c|c|c|}\hline
\multicolumn{1}{|c|}{\multirow{3}{*}{Method}} & \multicolumn{10}{c|}{Environment \& Robot Model}\\ \cline{2-11}
  &\multicolumn{2}{|c|}{\multirow{1}{*}{2D Point}}& \multicolumn{2}{|c|}{\multirow{1}{*}{2D Rigid}}& \multicolumn{2}{|c|}{\multirow{1}{*}{2D 2-link}}& \multicolumn{2}{|c|}{\multirow{1}{*}{2D 3-link}}& \multicolumn{2}{|c|}{\multirow{1}{*}{3D Point}}\\ \cline{2-11}
& seen & unseen& seen& unseen& seen& unseen& seen& unseen& seen& unseen\\ \hline
MPNet(NR) & 0.81&0.79              & 0.777&0.736            & 0.524&0.542                   & 0.380&0.386                   &0.762&0.753 \\\hline
MNP(Origin)           & 0.967&0.966            & 0.829&0.833            & 0.922&0.94                    & 0.933&0.947                   &0.872&0.891 \\\hline
MPNet(HR) & \textbf{1}&0.99        & 0.995&0.997            & 0.997&0.997                   & 0.979&0.954                   &0.999&0.999 \\\hline
MNP(RRT)  & \textbf{1}&\textbf{1}  & 0.998&0.997            & 0.998&\textbf{1}              & 0.994&0.991                   &0.999&\textbf{1} \\\hline
IRRT*     & \textbf{1}&\textbf{1}  & 0.998&0.994            & 0.994&0.998                   & 0.977&0.983                   &0.999&\textbf{1}  \\\hline
BIT*      & \textbf{1}&\textbf{1}  & \textbf{1}&\textbf{1}  & \textbf{1}&\textbf{1}         & \textbf{0.999}&\textbf{0.999} &\textbf{1}&\textbf{1}   \\\hline
\end{tabular}
\caption{Mean success rate comparison results. MPNet(NR) is MPNet with neuro replanning and MPNet(HR) is MPNet with hybrid replanning. MNP(Origin) has no replanner and MNP(RRT) has RRT as its replanner. For the top two methods of the table, since RRT is not used for replanning, they directly show the performance of the learned networks.}
\label{tab:suc}
\end{table*}
\end{small}

\begin{small}
    \begin{table*}[!ht]
\center
\begin{tabular}{|c|c|c|c|c|c|c|c|c|c|c|}\hline
\multicolumn{1}{|c|}{\multirow{3}{*}{Method}} & \multicolumn{10}{c|}{Environment \& Robot Model}\\ \cline{2-11}
  &\multicolumn{2}{|c|}{\multirow{1}{*}{2D Point}}& \multicolumn{2}{|c|}{\multirow{1}{*}{2D Rigid}}& \multicolumn{2}{|c|}{\multirow{1}{*}{2D 2-link}}& \multicolumn{2}{|c|}{\multirow{1}{*}{2D 3-link}}& \multicolumn{2}{|c|}{\multirow{1}{*}{3D Point}}\\ \cline{2-11}
& seen & unseen& seen& unseen& seen& unseen& seen& unseen& seen& unseen\\ \hline
MPNet(NR) & 0.025&0.026     & 0.0342&0.0301 & 0.03608&0.0336 & 0.0731&0.0747  &0.0167&0.0211  \\\hline
MNP(Origin)       & 0.0058&0.0069   & 0.0158&0.0168 & 0.0106&0.0105  & 0.0167&0.0162  &0.00989&0.0104\\\hline
MPNet(HR) & 0.029&0.033     & 0.0608&0.0552 & 0.0568&0.0530   & 0.238&0.194    &0.0267&0.0303\\\hline
MNP(RRT)  & \textbf{0.0066}&\textbf{0.0075}  & \textbf{0.0276}&\textbf{0.0318} & \textbf{0.0169}&\textbf{0.0127}  & \textbf{0.0555}&\textbf{0.0339} &\textbf{0.0156}&\textbf{0.0153}   \\\hline
IRRT*     & 0.116&0.147     & 0.345&0.211   & 0.207&0.172   & 0.565&0.542    &1.556&1.622\\\hline
BIT*      & 0.0395&0.0565   & 0.377&0.134   & 0.201&0.186   & 1.082&1.129    &0.980&0.946\\\hline
\end{tabular}
\caption{Mean computation time comparison results. Methods are compared in seen and unseen environments. Informed-RRT* and BIT* stop when the path length is within $110\%$ of MNP(RRT)'s. Note that MPNet(NR) and MNP have low planning success rates and their computation time are of limited significance, and we focus on the comparison of the four remaining methods.}
\label{tab:time}
\end{table*}
\end{small}


\par In this subsection, we compare the proposed MNP, MPNet and advanced sampling based methods IRRT* and BIT* in 2D and 3D environments with different robot models including the point mass, a rigid body, a two-link manipulator and a three-link manipulator. To avoid complete folding, the angle range of the link manipulators is set to $(-0.75\pi, 0.75\pi)$. Both seen and unseen environments will be used to assess how well the learned results generalize. For each of 900 seen environments and 100 unseen environments, we generate 10 random initial and goal configurations for comparison. Fig.~\ref{fig:multi strcuture} shows the paths planned by MNP with different robot models. Table~\ref{tab:time} and Table~\ref{tab:suc} show the computation time and success rate, respectively. The comparison of the paths length is shown in the appendix. 

From Table~\ref{tab:suc} we can see that without RRT as the replanner, the success rate of MPNet(NR) decreases rapidly as the dimension of the configuration space increases, while MNP(Origin) retains a high success rate. For the 2D 3-link manipulator where $C \subset \mathbb{R}^{5}$, the success rate of MNP(Origin) reaches $93.3\%$ while MPNet(NR) only has $38\%$. Table~\ref{tab:time} shows that MNP(RRT) takes the minimum computation time for planning compared with MPNet(HR), IRRT* and BIT*. For the 2D point mass, MPN(RRT) is $4$x faster than MPNet(HR) and $6$x faster than BIT* with a $100\%$ success rate. For the 2D rigid body, MPN(RRT) is $2$x faster than MPNet(HR) and $13$x faster than BIT* with a $99.8\%$ success rate. For the 2D 3-link manipulator, MPN(RRT) is $3$x faster than MPNet(HR) and $20$x faster than BIT* with a $99.4\%$ success rate. Comparison results in unseen environments also show that MPN(RRT) can generalize to new environments. With the increase of the dimension of the configuration space, the performance improvement of MNP(RRT) with respect to MPNet and BIT* becomes more significant. Note that, for training MNP and MPNet with the 2-link and 3-link manipulators, we remove the trivial scenarios where initial and goal configurations can be connected directly using a straight line from the training dataset, which seems to be helpful for  improving the performance of MNP.

\subsection{Comparisons using a 7-DOF Panda Arm}
\label{sec:comparison panda arm}
\begin{figure*}[htbp]
\centering
\hspace{-0.4cm}
\subfigure[Initial.]{
\includegraphics[width=1.1in]{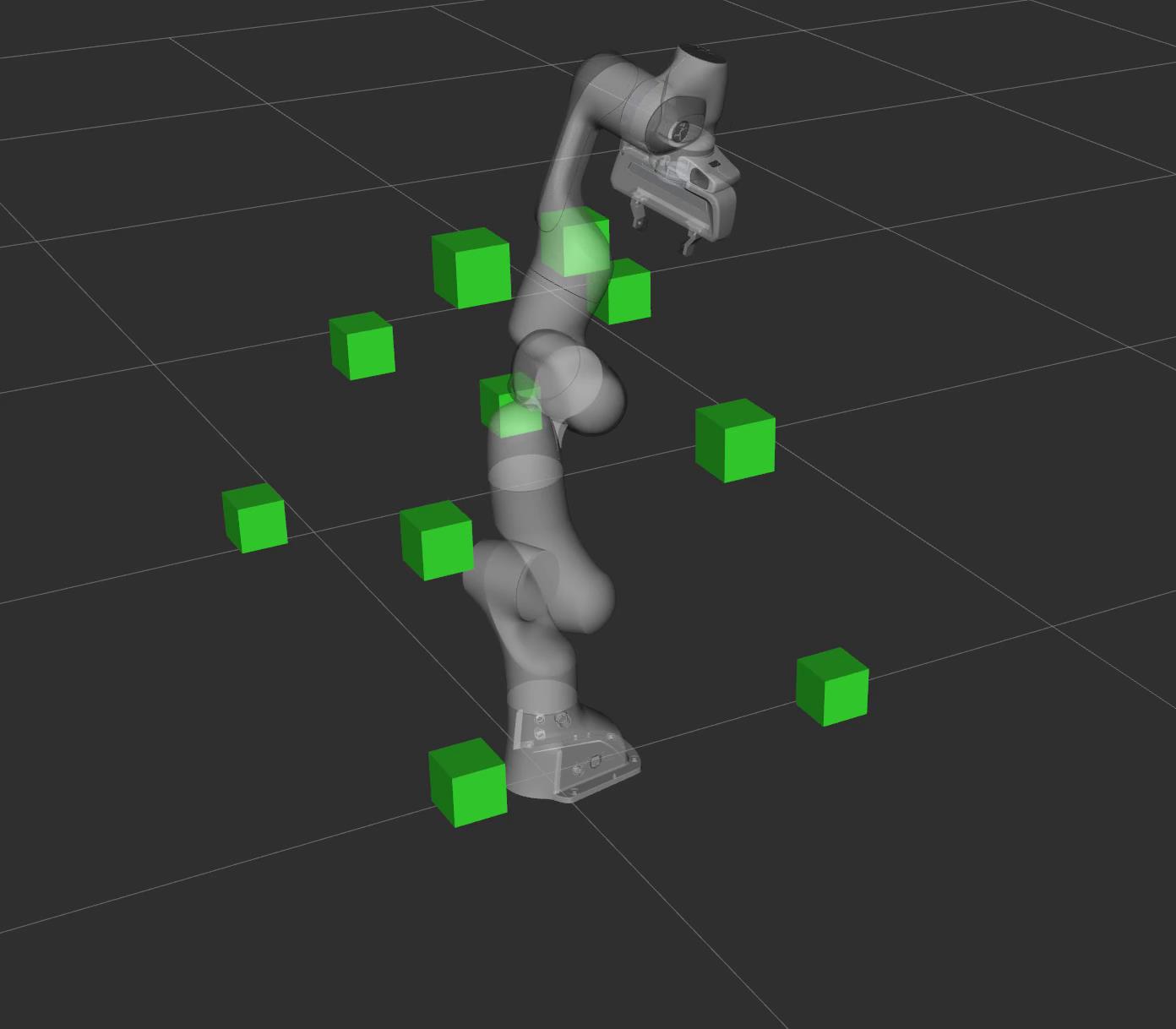}
}
\hspace{-0.4cm}
\subfigure[Step 1.]{
\centering
\includegraphics[width=1.1in]{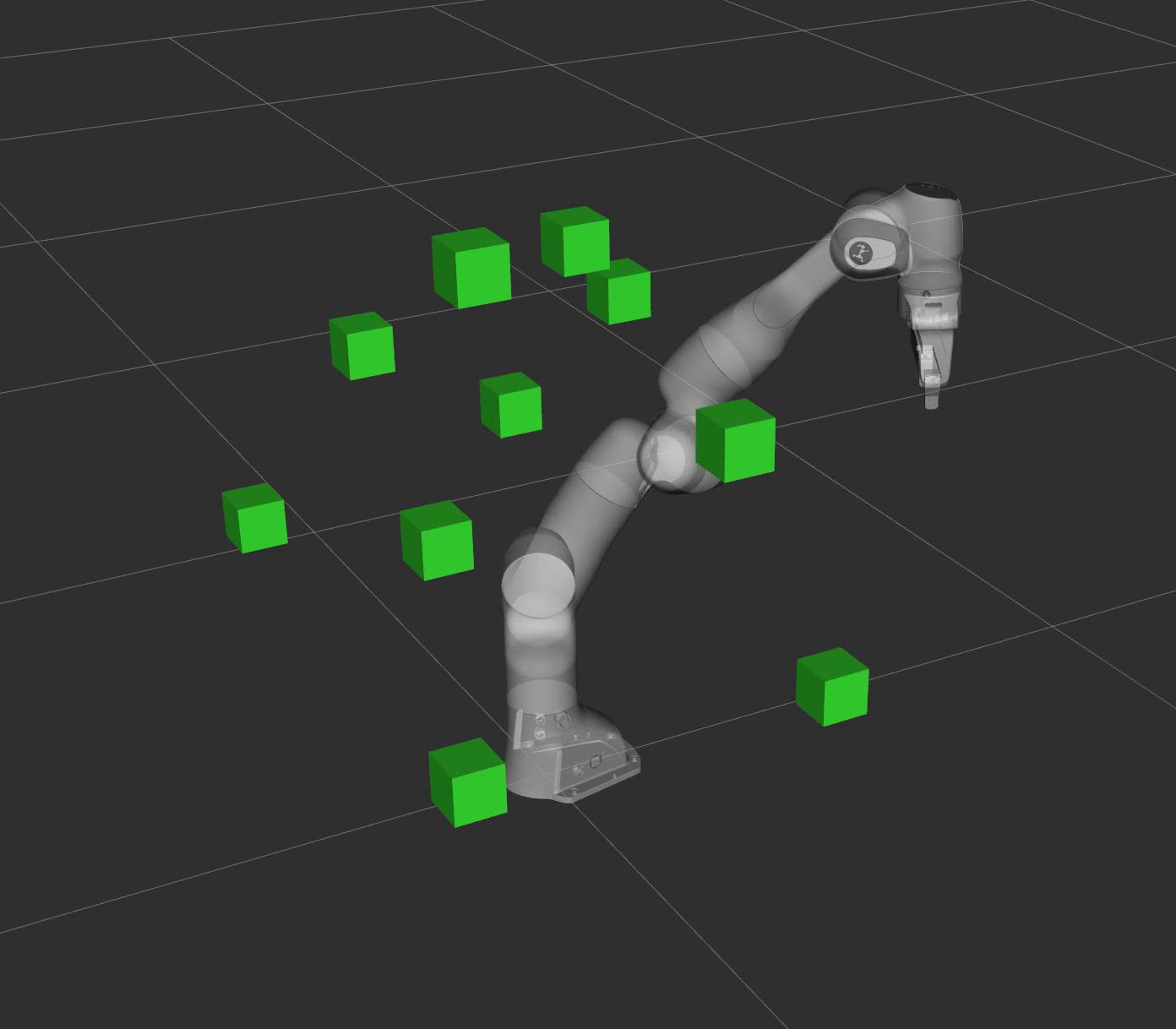}
}
\hspace{-0.4cm}
\subfigure[Step 2.]{
\centering
\includegraphics[width=1.1in]{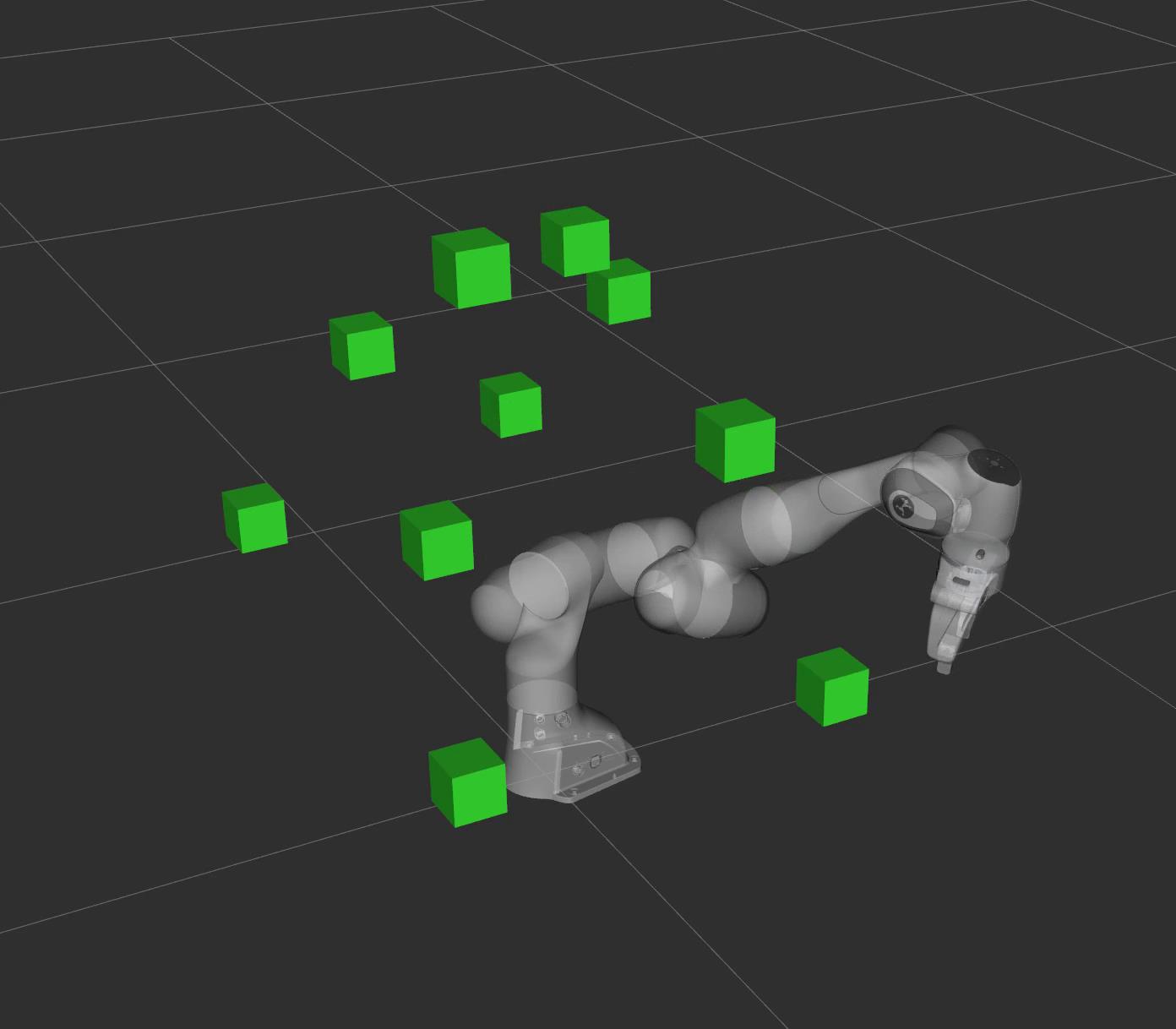}
}
\hspace{-0.4cm}
\subfigure[Step 3.]{
\centering
\includegraphics[width=1.1in]{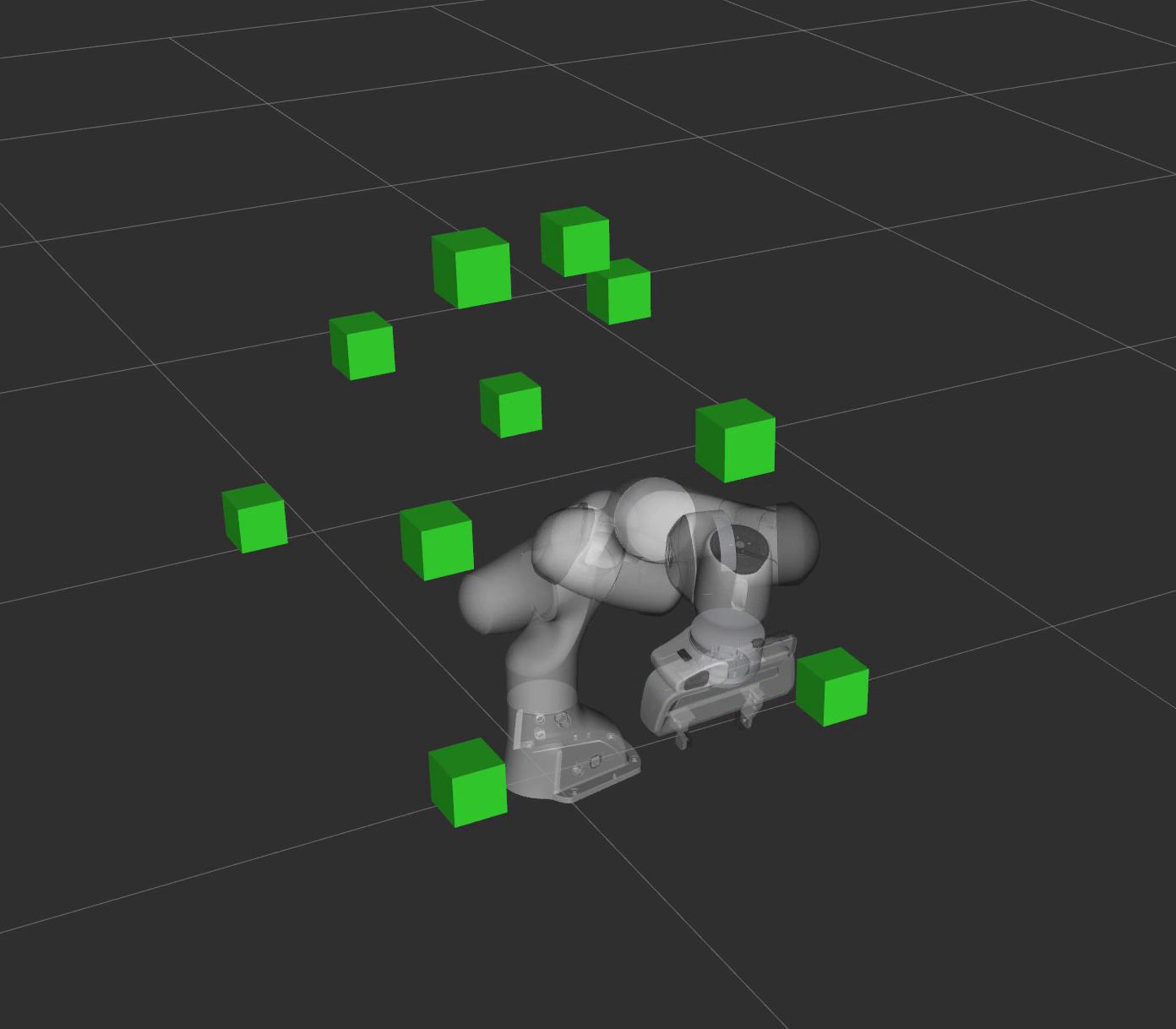}
}
\hspace{-0.4cm}
\subfigure[Goal.]{
\centering
\includegraphics[width=1.1in]{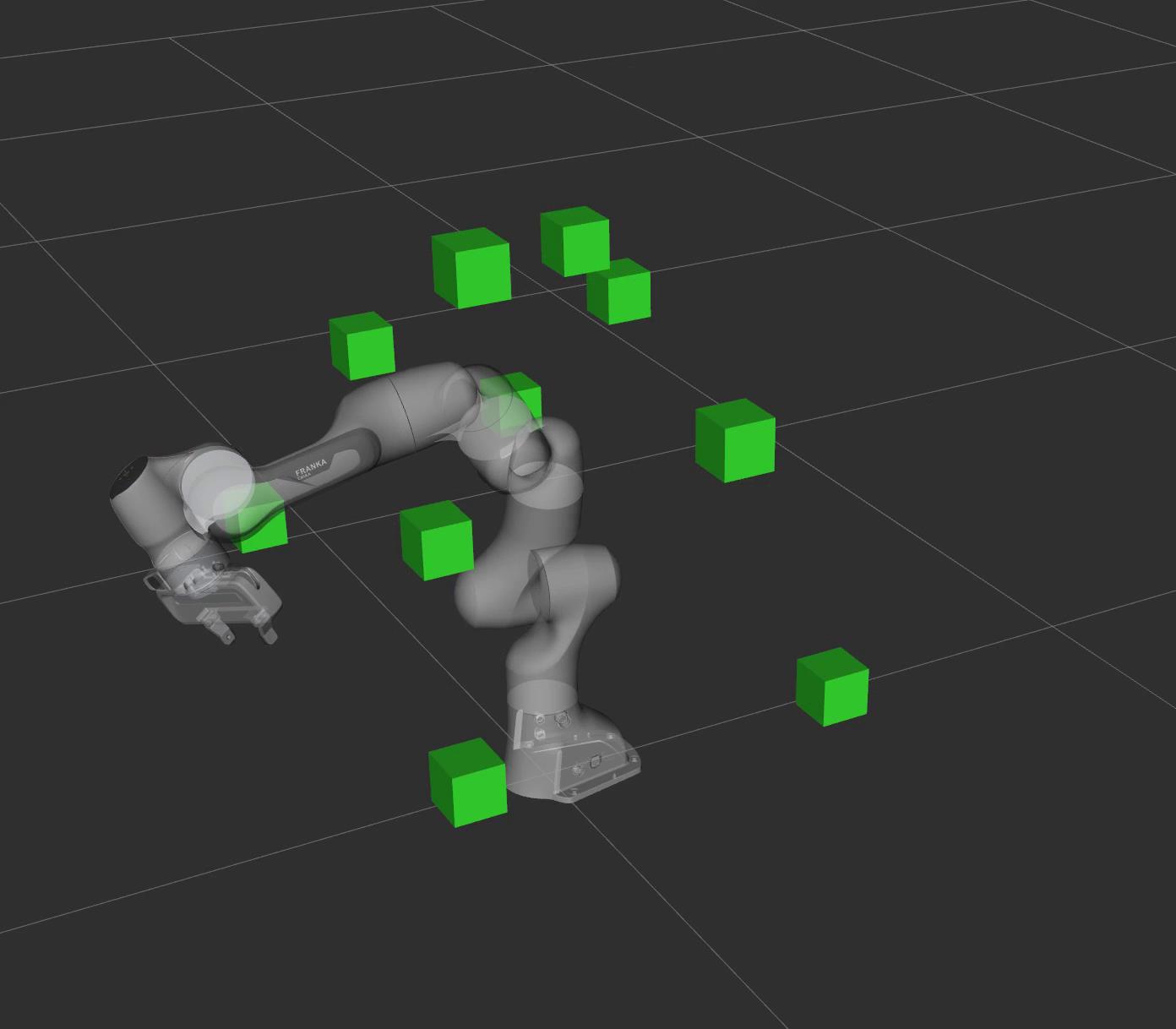}
}
\caption{Paths planned by MNP for Panda Arm where green cubes are obstacles.}
\label{fig:arm}
\end{figure*}

\begin{table}[!ht]
\center
\begin{tabular}{|c|c|c|c|}\hline
\multicolumn{1}{|c|}{\multirow{2}{*}{Method}} & \multicolumn{3}{c|}{Metrics}\\ \cline{2-4}
                & Time           & Length          & Suc Rate\\ \hline
MPNet(HR)       & 2.67           & \textbf{0.954}  & 0.897  \\ \hline
MNP(RRT)        & \textbf{1.56}  & 0.962           & 0.967 \\ \hline
BIT*($40\%$)      & 5.55           & 1.125           & 0.97 \\ \hline
BIT*($10\%$)      & 16.55          & \textbf{1}      & \textbf{0.983} \\ \hline

\end{tabular}
\caption{Comparison results with 7 DOF Panda Arm. BIT*($40\%$) means it stops when the path length is within $140\%$ of MNP(RRT)'s, and the same for BIT*($10\%$). The maximum planning time is set as $60$s for BIT*. For length comparison, the length of BIT*($10\%$) is set as the baseline.  }
\label{tab:arm}
\end{table}


\par We further evaluate MNP's performance in high dimensions via comparisons involving a 7-DOF Panda Arm. We evaluate the methods in $10$ random environments with $30$ random initial and goal configurations for each environments which are not in training dataset. Fig.~\ref{fig:arm} shows a trajectory planned by MNP(RRT) that Panda Arm follows from the initial to the goal configuration. Table~\ref{tab:arm} shows the comparison results.  
The success rate of MNP(RRT) is $96.7\%$, which is as high as BIT* whose maximum computation time is set to $60$s. MNP(RRT) achieves minimum computation time which is $1.7$x faster than MPNet(HR) and $11$x faster than BIT*($10\%$).

\par The overall comparison results show that our proposed MNP can learn and generate multiple solutions and it takes less computation time than MPNet and advanced classical sampling based methods such as IRRT* and BIT* to obtain a near-optimal solution. MNP can also generalize to new environments thanks to the Enet's ability to process point clouds. Finally, our method is effective for planning in high dimensional configuration space such as in robotic arm motion planning problems.
\section{Discussion}
\par In this paper, we address the multimodal property of motion planning problems and its influence on motion planning algorithms based on deep imitation learning, and we propose to use the mixture Gaussian model with mixture density networks to learn from multimodal solutions. In MPNet, the authors use dropout layers to introduce stochasticity, but their aim is to generate different configurations for replanning, and they do not focus on learning multimodal solutions. Although the dropout mechanism is proven to be a Bayesian approximation and has the potential to learn multimodal solutions \cite{2016_ICML_Gal_dropout_as_bayesian}, its ability to learn multimodal solutions degenerates rapidly as the size of training dataset increases. For the MPNet where the 
dropout probability is set to be $0.5$ for each layer, it demonstrates strong ability to learn multimodal solutions when the size of the training dataset is smaller than $10$k. However, its performance degrades a lot when the size of the dataset increases to $100$k and $1000$k. When the dataset grows lager, MPNet seems to learn the "average" of the multimodal solution as shown in Sec~\ref{sec:multimodal comparison} and Fig.~\ref{fig:multimodal comparison}, which results in a high collision rate of the predicted configurations and low efficiency of the algorithm. Increasing the complexity of the network may improve the effectiveness of the dropout mechanism, but the network may become very inefficient computationally. For example, the Pnet of 
 MPNet has $3.76$M parameters, but it still fails to learn multimodal solutions with the point mass model in 2D environments. 


\section{Conclusion}
\label{sec:conclusion}
\par In this paper, we proposed a learning-based motion planning method termed Multimodal Neuron Planner (MNP). MPN is comprised of an environment encoding network (Enet) and a multimodal planning network (Pnet). We utilize the mixture Gaussian model to describe the distribution of the configuration, which enables Pnet to learn and predict multiple solutions. Simulation results verified MNP's ability to learn and produce multiple solutions and demonstrated that MNP outperforms the SOTA learning based method MPNet and advanced sampling based method IRRT* and BIT*.

\appendix
\subsection{Online Planning Algorithm}
\par After offline training, we can plan a path for a motion planning problem online with the trained Enet and Pnet models. The key idea is to generate waypoints iteratively by Pnet and finally to form a path. To make the algorithm complete and more efficient, the online planning algorithm utilizes additional mechanisms such as bi-directional planning and replanning. The outline of the online planning algorithm is similar to MPNet with some improvements.

\subsection{Bi-directional planner}
\label{sec:bi_dirctional planner}

\begin{algorithm}
    \caption{Bi-directionalPlanner($c_{\mathrm{init}}, c_{\mathrm{goal}}, Z$)}
    \textbf{Input}: Initial and goal configuration$c_{\mathrm{init}}, c_{\mathrm{goal}}$, environment encoding $Z$.\\
    \textbf{Output}: Solution path $\tau$.\\
    $\tau^a \gets \left \{ c_{\mathrm{init}}  \right \} $, $\tau^b \gets \left \{ c_{\mathrm{goal}}  \right \} $;\\
    \For{$i \gets 0$ \textbf{to} $N_{\mathrm{iter}}$}
    {
        $(\alpha, \mu, \sigma) \gets Pnet(\tau^a_{\mathrm{end}}, \tau^b_{\mathrm{end}}, Z)$\\
        \For{$j \gets 0$ \textbf{to} $N_{\mathrm{col}}$}
        {   \label{alg1:ck_beg}
            $c_{\mathrm{new}} \gets $ Sample$(p(c|\alpha, \mu, \sigma))$\\
            \If{$\mathrm{SteerTo}(\tau^a_{\mathrm{end}}, c_{\mathrm{new}})$}
            {
                $\tau^a \gets \tau^a \cup \{c_{\mathrm{new}}$\}\\
                break
            }
        }
        \If{$j\ge N_{\mathrm{col}}$}
        {
            $\tau^a \gets \tau^a \cup \{c_{\mathrm{new}}$\}\\
        }
        \label{alg1:ck_end}
        \If{$\mathrm{SteerTo}(\tau^a_{\mathrm{end}}, \tau^b_{\mathrm{end}})$}
        {   \label{alg1:str_beg}
            $\tau \gets \mathrm{Concatenate} (\tau^a, \tau^b)$ \\
            \Return {$\tau$} \\
        }\label{alg1:str_end}
        SWAP($\tau_a, \tau_b$)\\ \label{alg1:swap}
    }
    \Return {$\emptyset$} \\
    \label{alg:bi_dirctional planner}
\end{algorithm}

\par Bi-directional planning is utilized to improve the efficiency of the planning algorithm and is widely adopted, e.g., in RRT-connect \cite{2000_ICRA_Kuffner_RRTConnect} and MPNet. The outline of the Bi-directional Planner (BP) using Pnet is shown in Algorithm 1. BP takes in the initial configuration $c_{\mathrm{init}}$, the goal configuration $c_{\mathrm{goal}}$ and the environment latent vector $Z$ obtained by Enet, and it generate two paths where the first path $\tau^a$ begins with $c_{\mathrm{init}}$ and the second path $\tau^b$ begins with $c_{\mathrm{goal}}$. The two paths attempt to approach each other incrementally by alternately adding new configurations. Specifically, in one iteration, $\tau^a$ is extended towards $\tau^b_{\mathrm{end}}$ with $\tau^a$'s last configuration $\tau^a_{\mathrm{end}}$ as the current configuration and $\tau^b$'s last configuration $\tau^b_{\mathrm{end}}$ as the goal configuration. A new configuration $c_{\mathrm{new}}$ is added to $\tau^a$ as the last configuration. In the next iteration, $\tau^b$ is extended towards $\tau^a_{\mathrm{end}}$ by swapping $\tau^a$ and $\tau^b$ (line \ref{alg1:swap}). To generate the new configuration $c_{\mathrm{new}}$, Pnet first generates the predicting distribution, and then $c_{\mathrm{new}}$ is sampled according to (3). Finally, collision detection is performed between $\tau^a_{\mathrm{end}}$ and $c_{\mathrm{new}}$ up to $N_{\mathrm{col}}$ times to check if we can connect them (line \ref{alg1:ck_beg}-\ref{alg1:ck_end}). The collision detection function SteerTo$(c_1, c_2)$ takes in two configurations and applies a linear interpolation to obtain intermediate configurations between them. It then checks if all intermediate configurations are in $C_{\mathrm{free}}$ through the collision check function $\Phi$ mentioned in Section~3. If the two paths $\tau^a$ and $\tau^b$ can be directly connected, then the algorithm returns the concatenated path (line \ref{alg1:str_beg}-\ref{alg1:str_end}).
\subsubsection{Replanning}
\label{sec:replanning}
\begin{algorithm}
    \caption{Replan($\tau, \mathrm{planner}$)}
    \textbf{Input}: Original path $\tau$, planner for replanning.\\
    \textbf{Output}: Replan path $\tau^{\mathrm{new}}$.\\
    $\tau^{\mathrm{new}} \gets \emptyset$\\
    $\tau \gets \mathrm{PathSimplify}(\tau)$\\ \label{alg2:simp}
    \For{$i \gets 1$ \textbf{to} $\tau.\mathrm{length()}$}
    {
        \If{$\mathrm{SteerTo}(\tau_i, \tau_{i+1})$}
        {   \label{alg2:feasible_beg}
            $\tau^{\mathrm{new}} \gets \tau^{\mathrm{new}} \cup \{\tau_{i+1}$\}\\
        }\label{alg2:feasible_end}
        \Else{
            \label{alg2:rp_beg}
            $\tau^{\mathrm{connect}} \gets \mathrm{planner}(\tau_i, \tau_{i+1})$\\
            $\tau^{\mathrm{connect}} \gets \mathrm{PathSimplify}(\tau^{\mathrm{connect}})$\\
            $\tau^{\mathrm{new}} \gets \tau^{\mathrm{new}} \cup \tau^{\mathrm{connect}}$\\
        }
        \label{alg2:rp_end}
    }

    \Return {$\tau^{\mathrm{new}}$} \\
    \label{alg:replan}
\end{algorithm}
\par The replanning mechanism is introduced to make the path complete and feasible, which checks if there is collision between all adjacent configurations and repairs the path if collision occurs. The outline of the replanning is shown in Algorithm 2, which takes a path and a planner as input and outputs a new path. At first, a PathSimplify operation (line \ref{alg2:simp}) is performed to remove all the colliding and redundant configurations in the path. For example, for a path $\tau = \{c_0,c_1,c_2,\dots,c_l\}$, if $c_0$ and $c_2$ could be connected directly without collision (SteerTo($c_0,c_2$) returns True), then $c_1$ will be removed to make the following replanning procedure more computationally efficient. For the rest of configurations, collision check between adjacent configurations is performed by SteerTo. If there is no collision, then the adjacent configurations will be added to the new path (line \ref{alg2:feasible_beg}-\ref{alg2:feasible_end}), otherwise a connection path $\tau^{\mathrm{connect}}$ will be planed by the input planner and added to the new path (line \ref{alg2:rp_beg}-\ref{alg2:rp_end}). MPNet uses BP or RRT as the Replan's planner, but in this paper we only use RRT. Through replanning with RRT the new path is expected to be complete and collision free.
\subsection{Muiltimodal neuron planner}
\label{sec:Muiltimodal neuro planner}
\begin{algorithm}
    \textbf{Input}: Initial and goal configuration $c_{\mathrm{init}}$ and $ c_{\mathrm{goal}}$, point clouds $X_{\mathrm{cloud}}$.\\
    \textbf{Output}: Solution path $\tau$.\\
    \caption{MuiltimodalNeuronPlanner($c_{\mathrm{init}}, c_{\mathrm{goal}}, X_{\mathrm{cloud}}$)}
    $ Z \gets Enet(X_{\mathrm{cloud}})$\\
    $\tau \gets $ Bi-directionalPlanner$(c_{\mathrm{init}}, c_{\mathrm{goal}}, Z)$\\
    $\tau \gets \mathrm{PathSimplify} (\tau)$\\
    \If{$\mathrm{IsFeasible}(\tau)$}
    {
        \Return{$\tau$}
    }
    \Else{
    $\tau \gets \mathrm{Replan}(\tau, \mathrm{RRT})$\\
    \Return{$\tau$}
    }

    \label{alg:Muiltimodal Neuro Planner}
\end{algorithm}
\par The outline of the whole proposed online motion planning algorithm Muiltimodal Neuron Planner~(MNP) is shown in Algorithm~3. It takes in the point clouds $X_{\mathrm{cloud}}$, initial configuration $c_{\mathrm{init}}$ and goal configuration $c_{\mathrm{goal}}$ and outputs the solution path. 
The environment latent vector $Z$ is firstly obtained 
through Enet. Then Bi-directionalPlanner is applied to get an initial solution path, which is simplified by PathSimplify.  After that, function IsFeasible($\tau$) is called to check if there is collision between any adjacent configurations in the path and returns True if no collision occurs. If the path is feasible, then MNP terminates and returns the path. Otherwise, Replan with RRT is performed to get a feasible path. Note that Replan with RRT ensures that MNP is a complete motion planner,  i.e., it finds a feasible solution if the solution exists.

\subsection{Implementation Details And Results}
\label{sec:implementation details}
\par The deep networks are implemented using Pytorch \cite{2019_NIPS_Paszke_pytorch} in Python. Since the comparisons are performed in the C++ environment, Torch Script is used to enable the use of Pytorch model in C++. We have three types of environment point clouds, and two of them are called Simple 2D and Complex 3D similar to MPNet \cite{2020_TRO_Qureshi_Motion_planning_networks}. The third point cloud used in the 7-DOF Panda Arm simulation is generated by our own. The Simple 2D and complex 3D are represented by $1400$ and $1000$ points, respectively. The environment of 7-DOF Panda Arm is represented by $500$ points, which are randomly sampled from the surface of obstacles. The training data is generated by RRT* and BIT*, and we use the standard OMPL~\cite{2012_IRAM_Sucan_ompl} C++ implementation for all sampling based methods. For training, in Simple 2D or Complex 3D cases, we generate $400$ paths for each of the $900$ environments. These $900$ environments are called seen environments. The testing is performed in $1000$ environments where $900$ of them are seen during the training and the other $100$ environments are randomly generated and thus called unseen. 
For comparisons, we generate $10$ random initial and goal configurations for each of the $1000$ environments for each robot model. 
For the training in the 7-DOF Panda Arm experiment, we generate $1000$ paths for each of the $10$ randomly generated environments. The testing is performed in these same environments with $30$ randomly generated initial and goal configurations for each of them. Note that in all comparison studies, trivial initial and goal configurations that can be directly connected via a straight line are discarded and replaced. We use NVIDIA RTX3090 for training and NVIDIA RTX3070 Laptop, i7-11800H for testing.

\begin{figure*}[htbp]
\centering
\subfigure[2D point.]{
\includegraphics[width=1.2in]{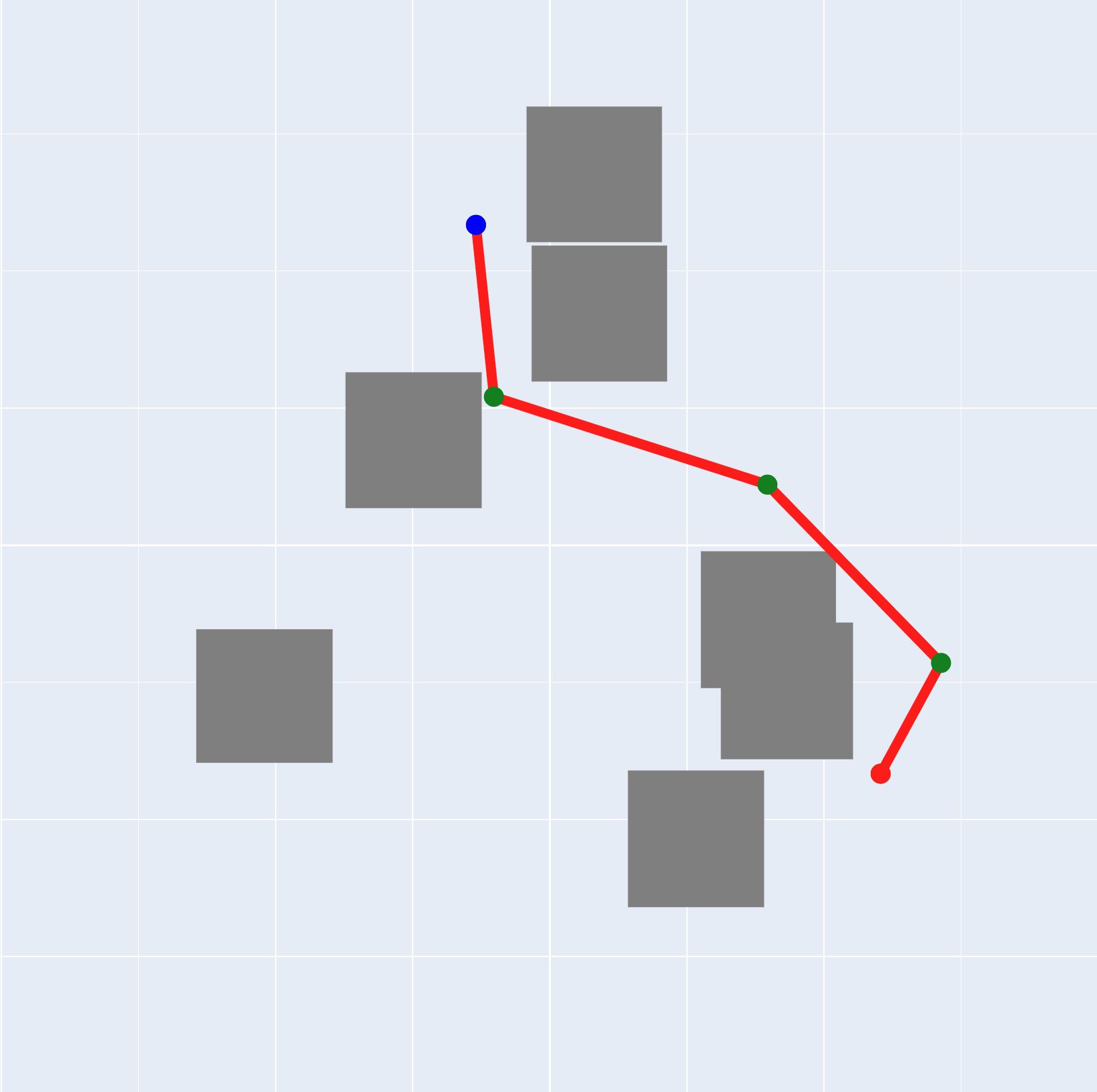}
}
\subfigure[2D rigid.]{
\includegraphics[width=1.2in]{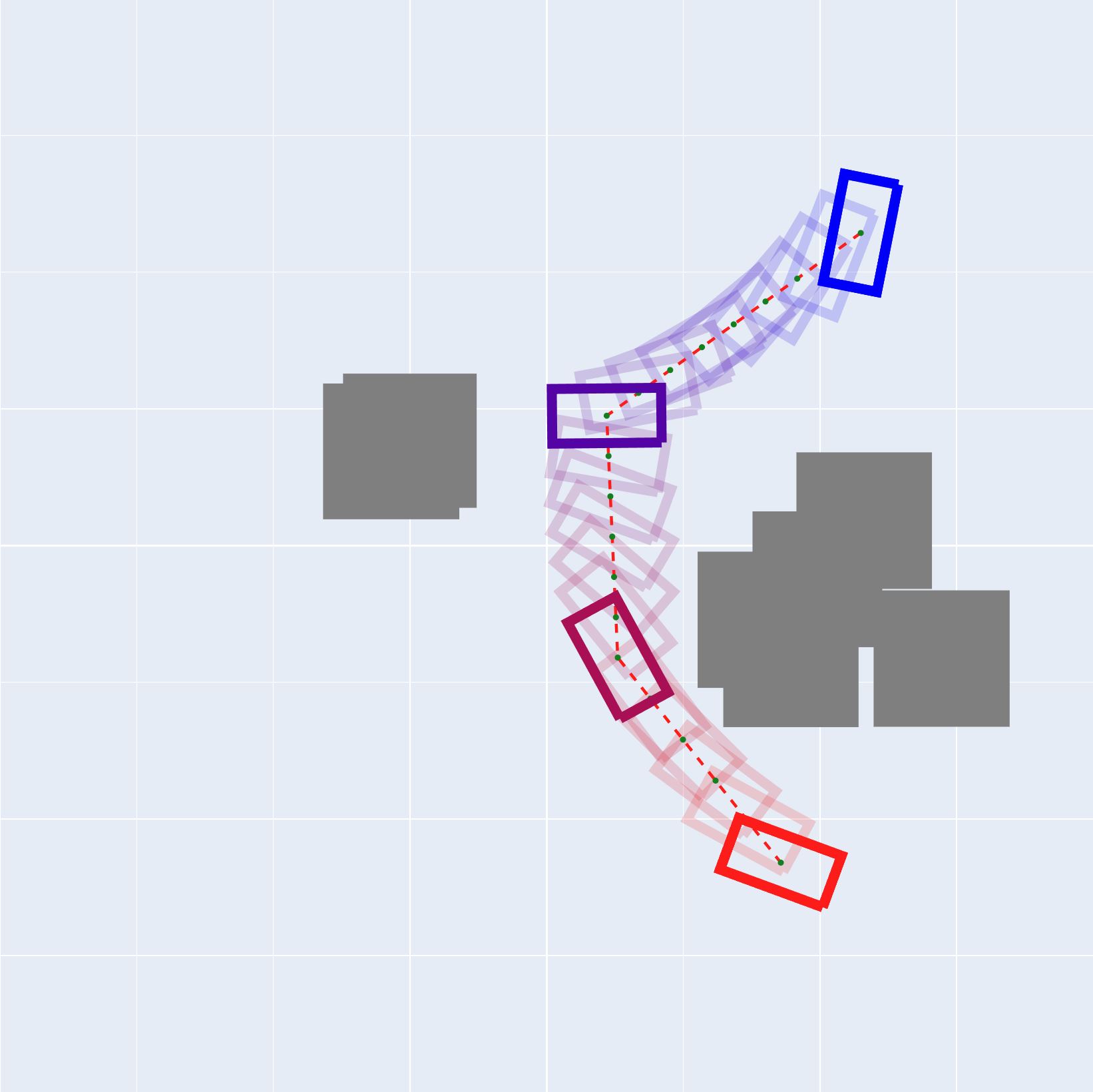}
}
\subfigure[2D 2-link.]{
\centering
\includegraphics[width=1.2in]{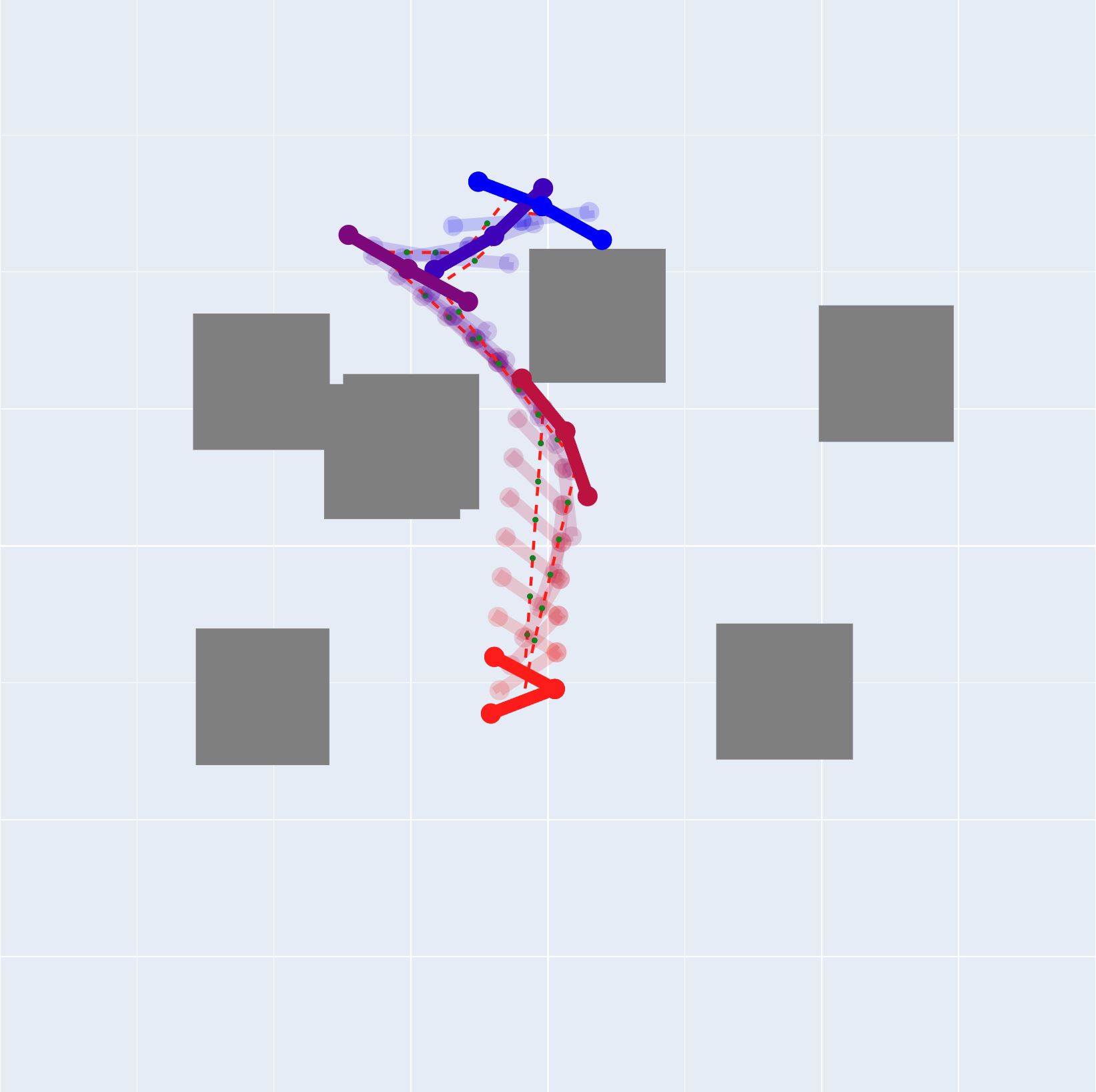}
}
\subfigure[2D 3-link.]{
\centering
\includegraphics[width=1.2in]{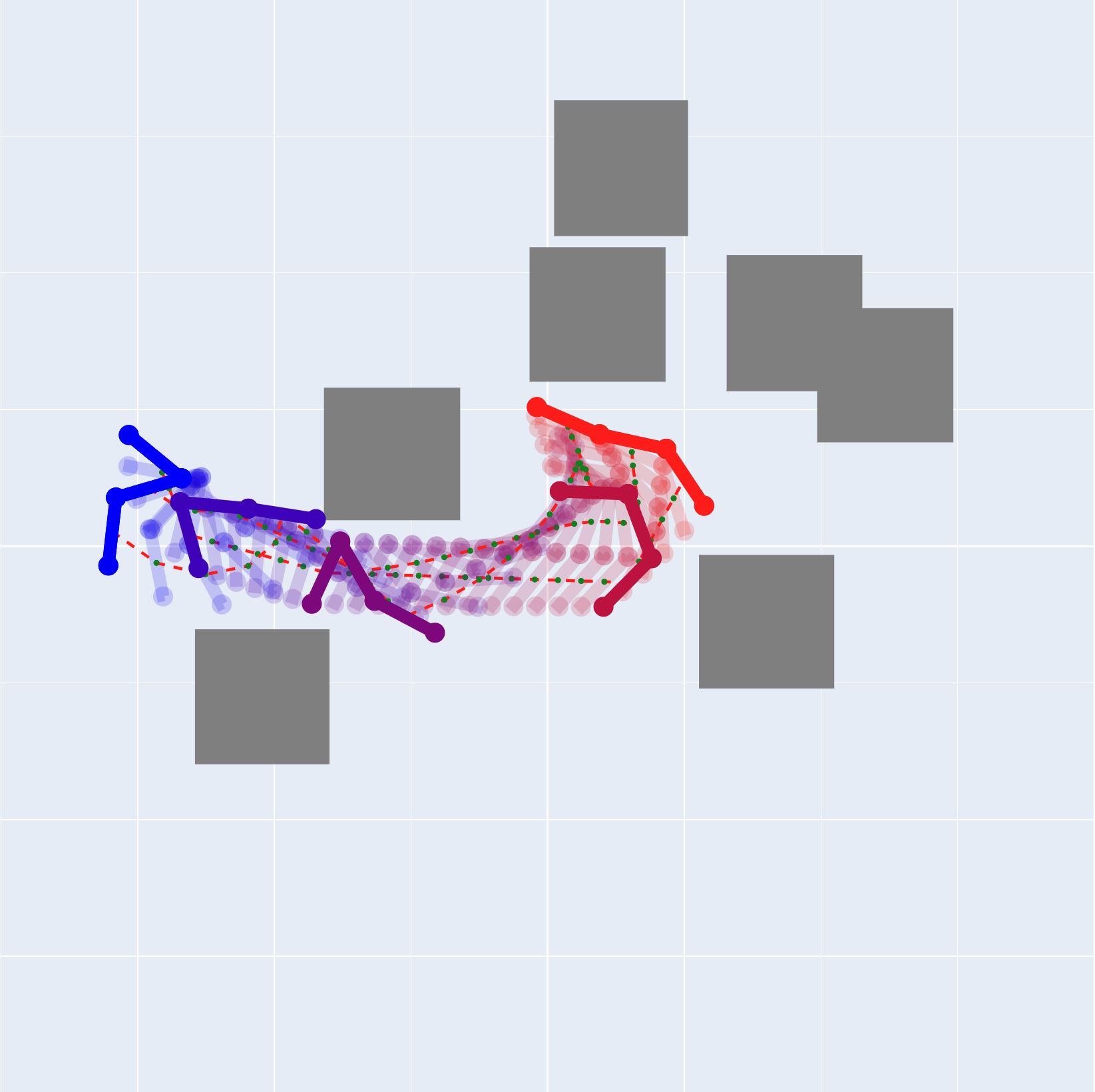}
}
\subfigure[3D point.]{
\includegraphics[width=1.2in]{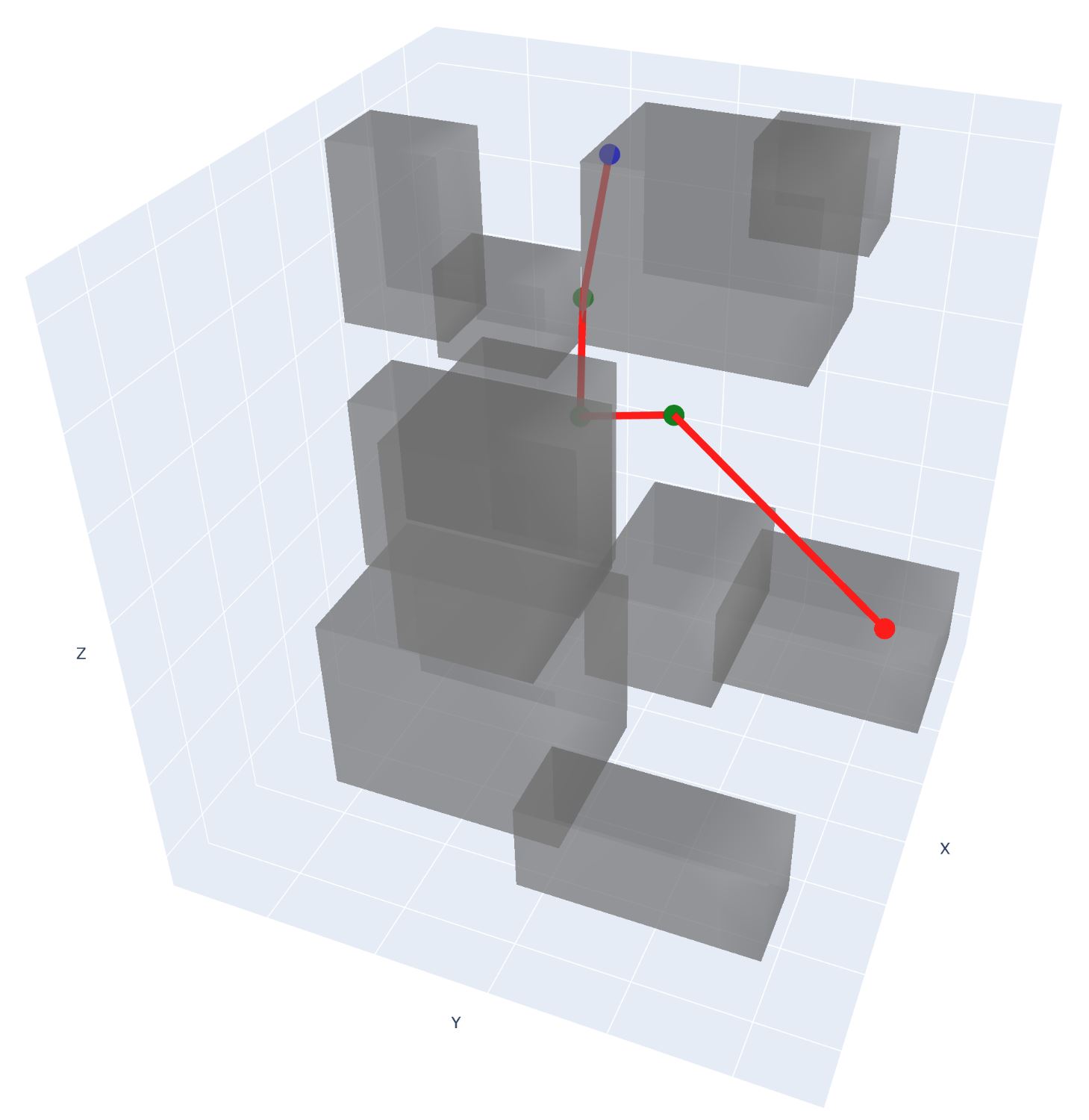}
}
\subfigure[2D point.]{
\includegraphics[width=1.2in]{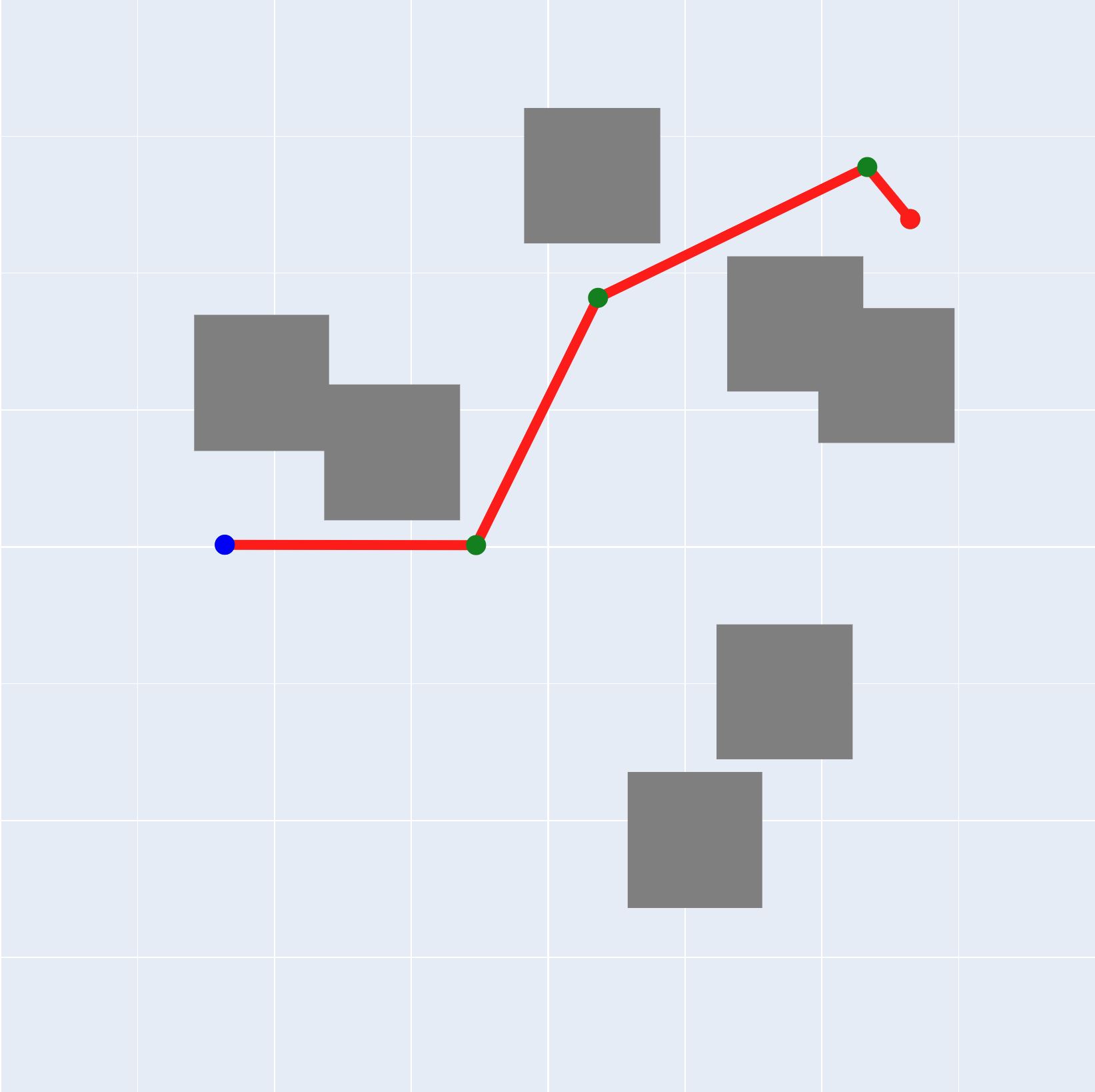}
}
\subfigure[2D rigid.]{
\includegraphics[width=1.2in]{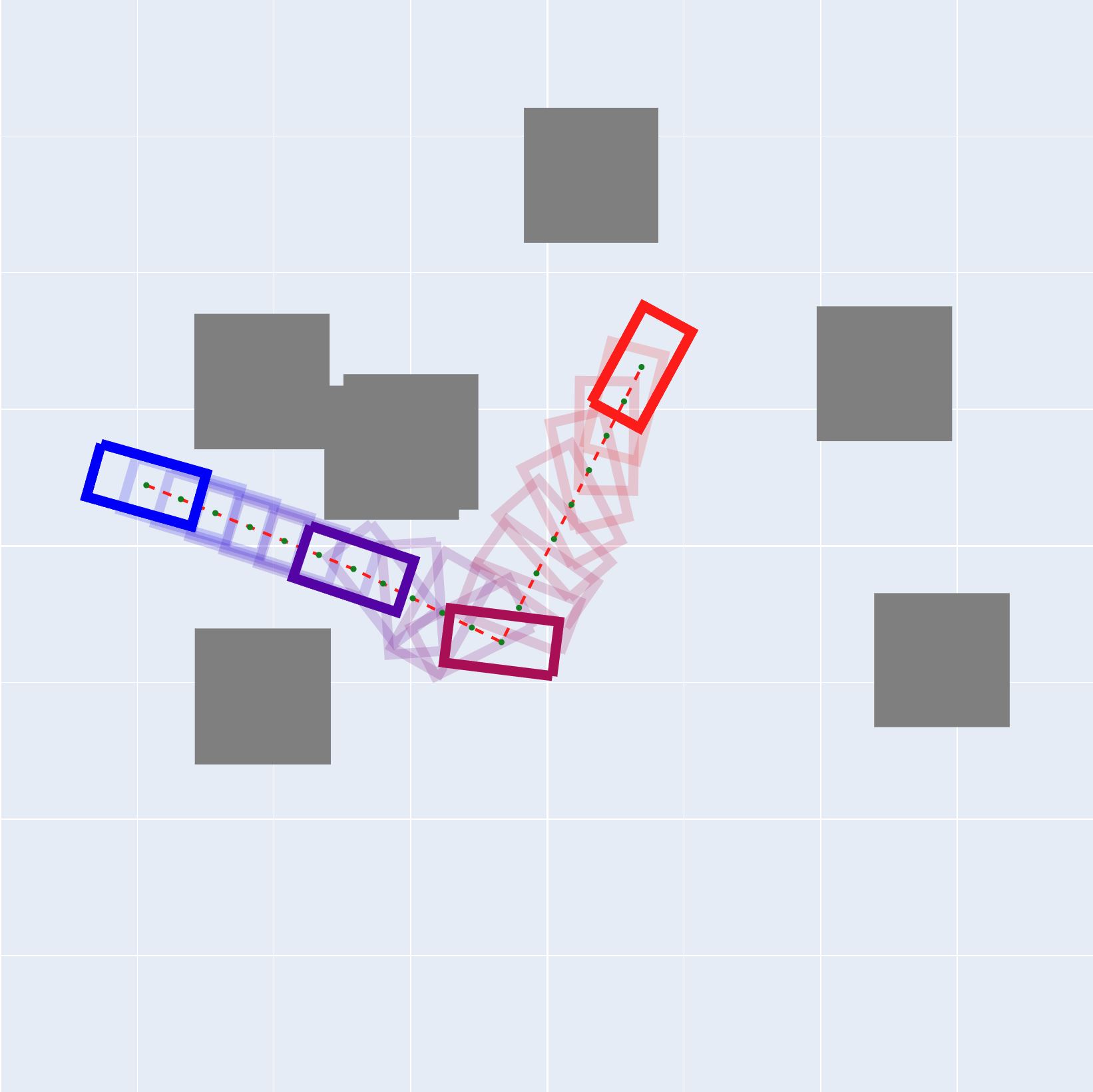}
}
\subfigure[2D 2-link.]{
\centering
\includegraphics[width=1.2in]{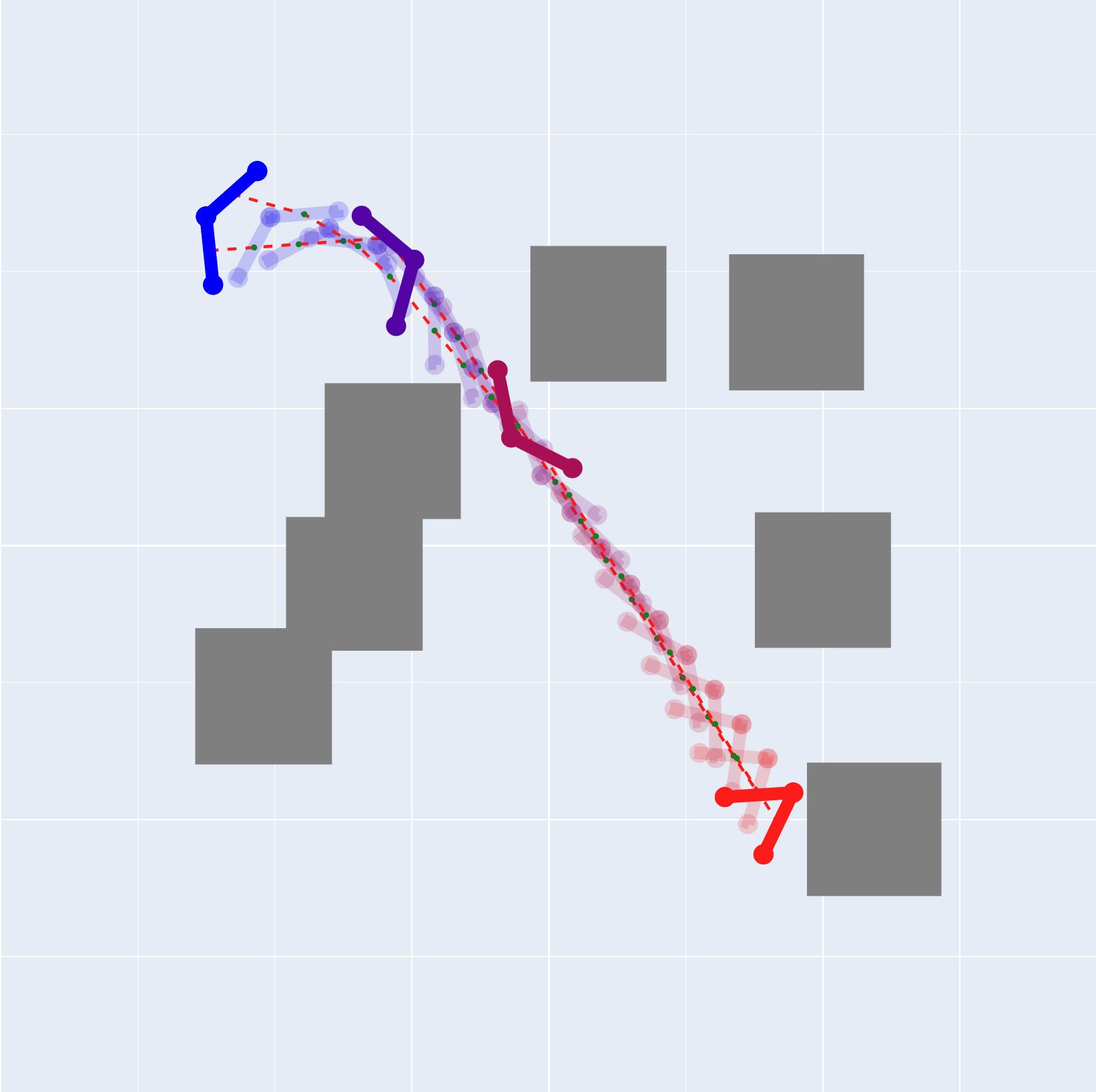}
}
\subfigure[2D 3-link.]{
\centering
\includegraphics[width=1.2in]{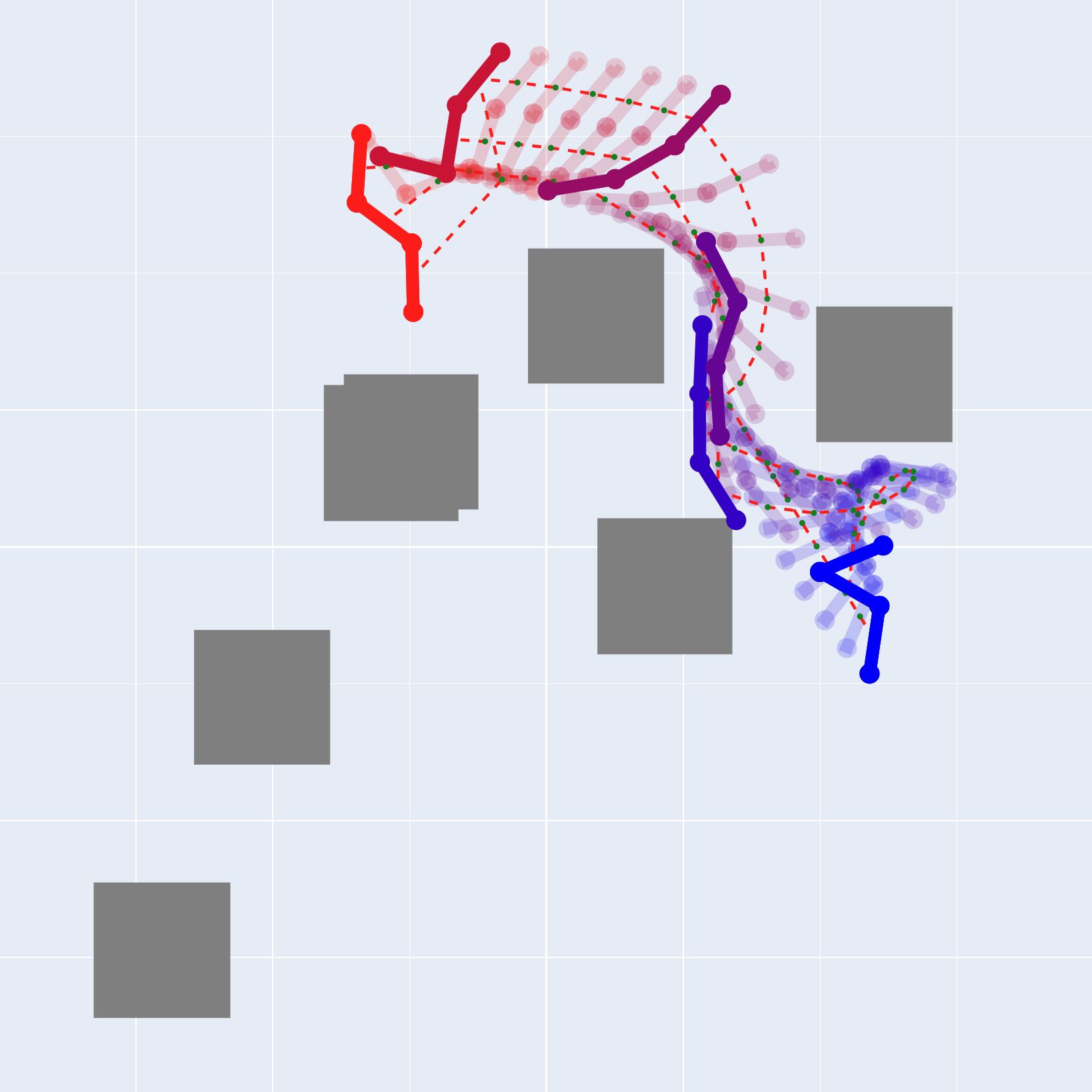}
}
\subfigure[3D point.]{
\includegraphics[width=1.2in]{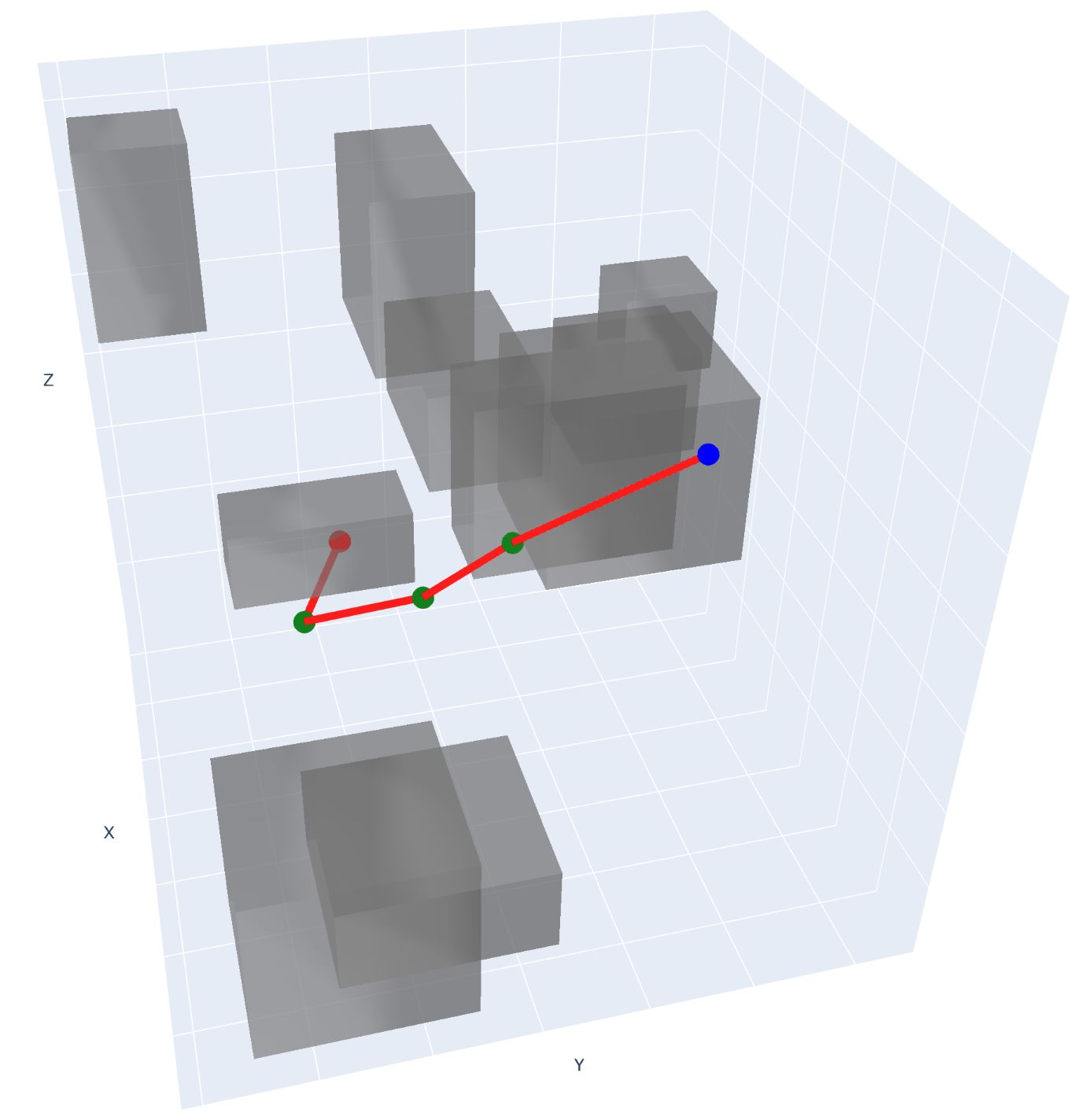}
}
\subfigure[2D point.]{
\includegraphics[width=1.2in]{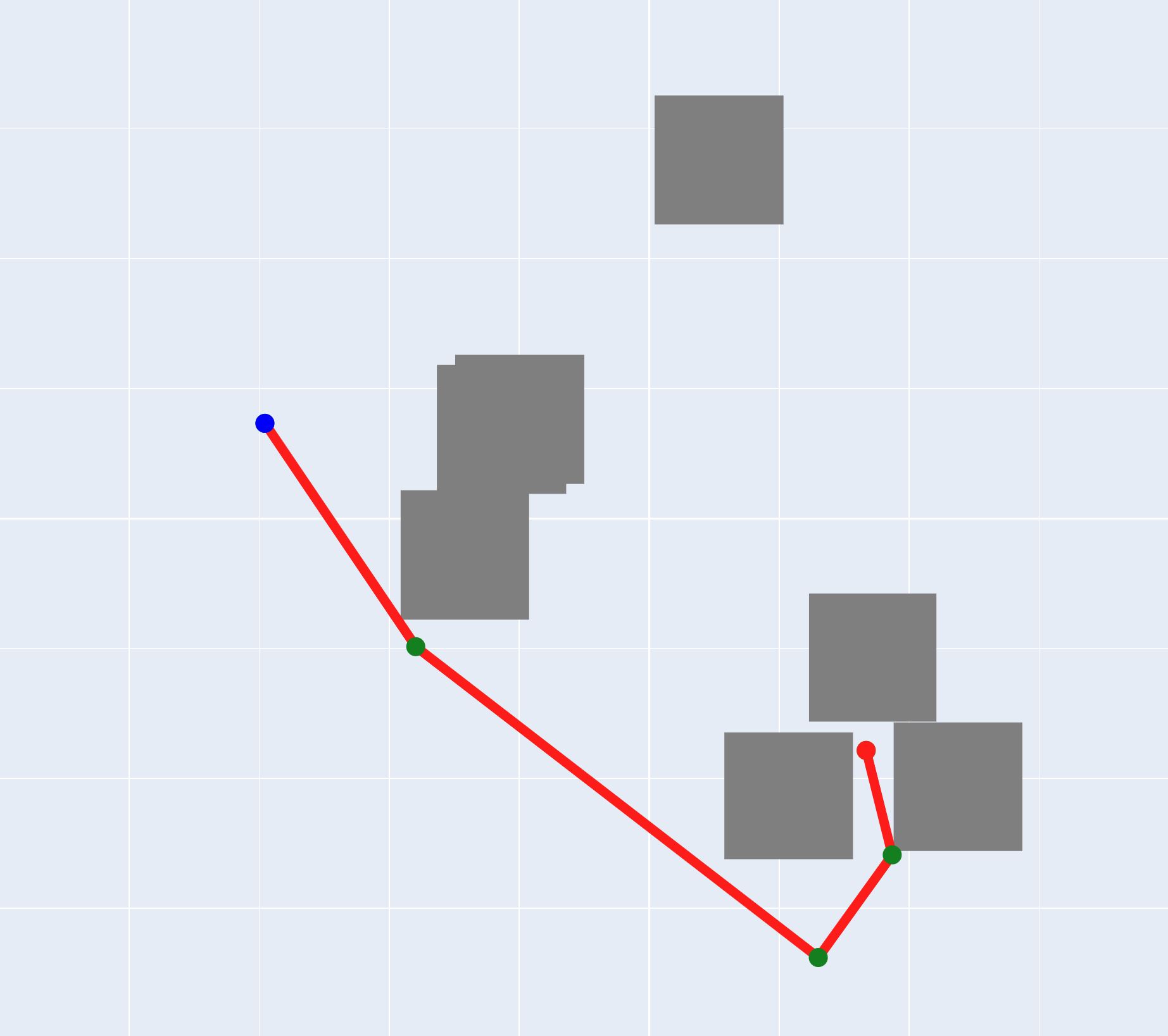}
}
\subfigure[2D rigid.]{
\includegraphics[width=1.2in]{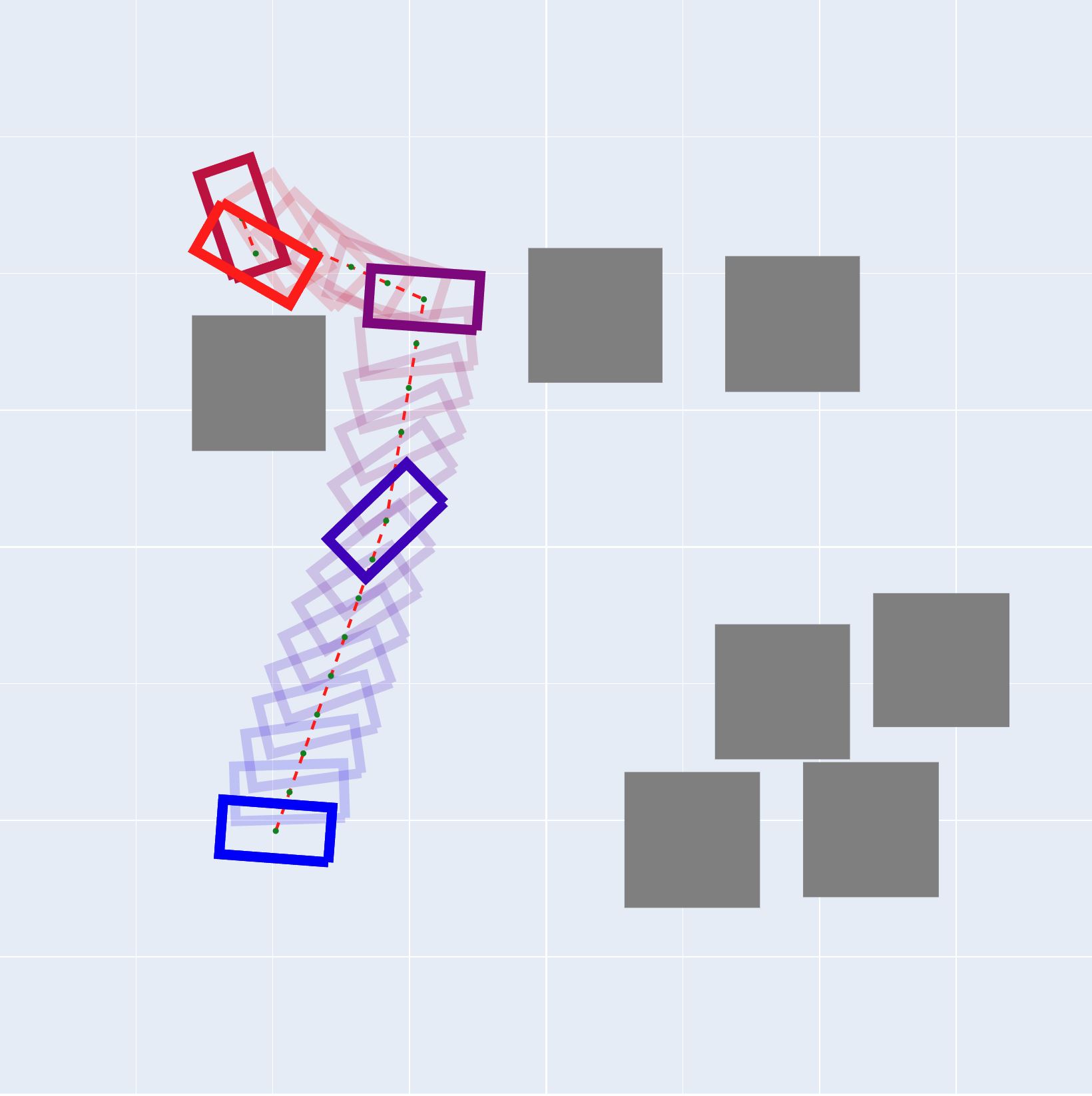}
}
\subfigure[2D 2-link.]{
\centering
\includegraphics[width=1.2in]{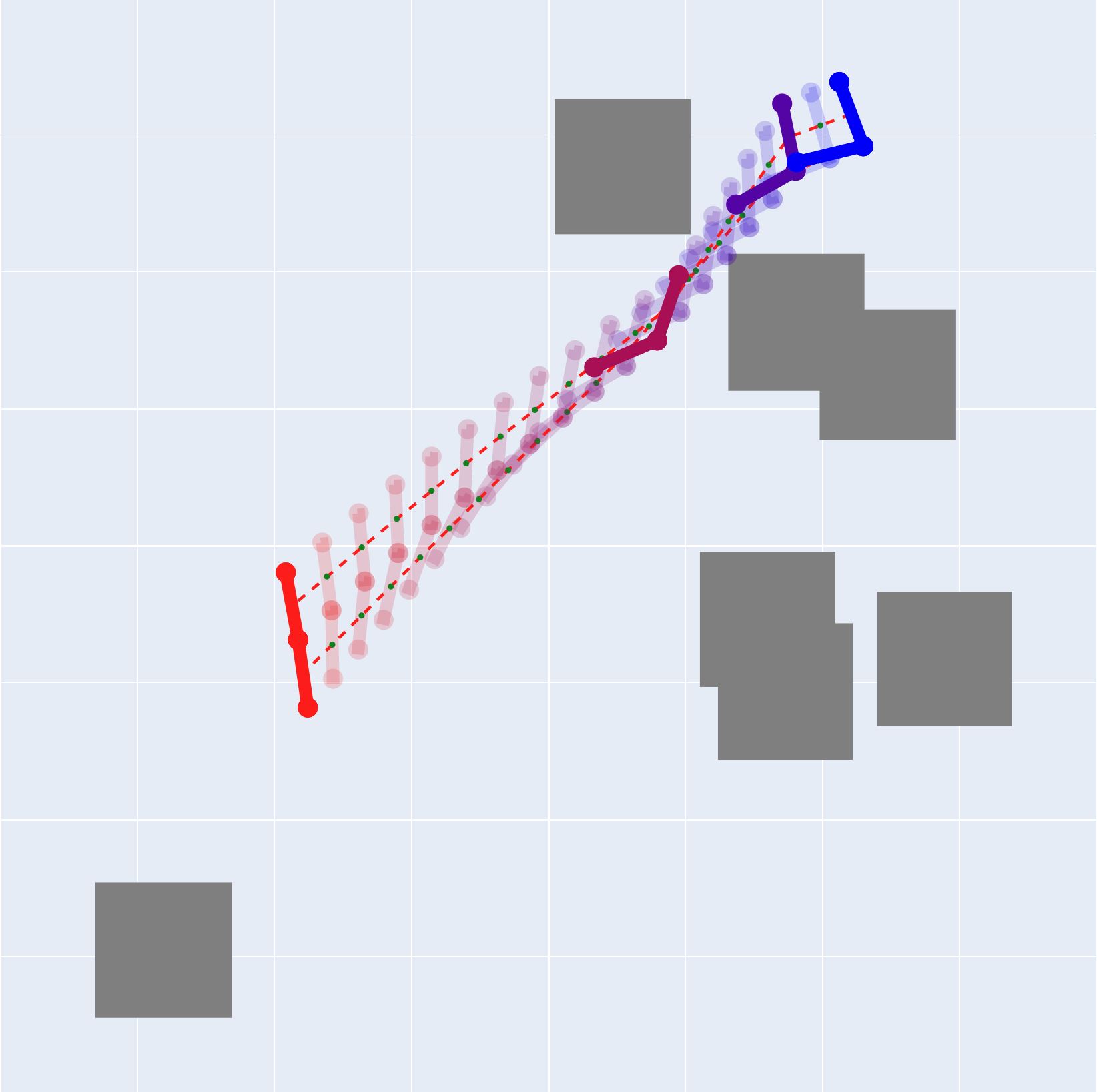}
}
\subfigure[2D 3-link.]{
\centering
\includegraphics[width=1.2in]{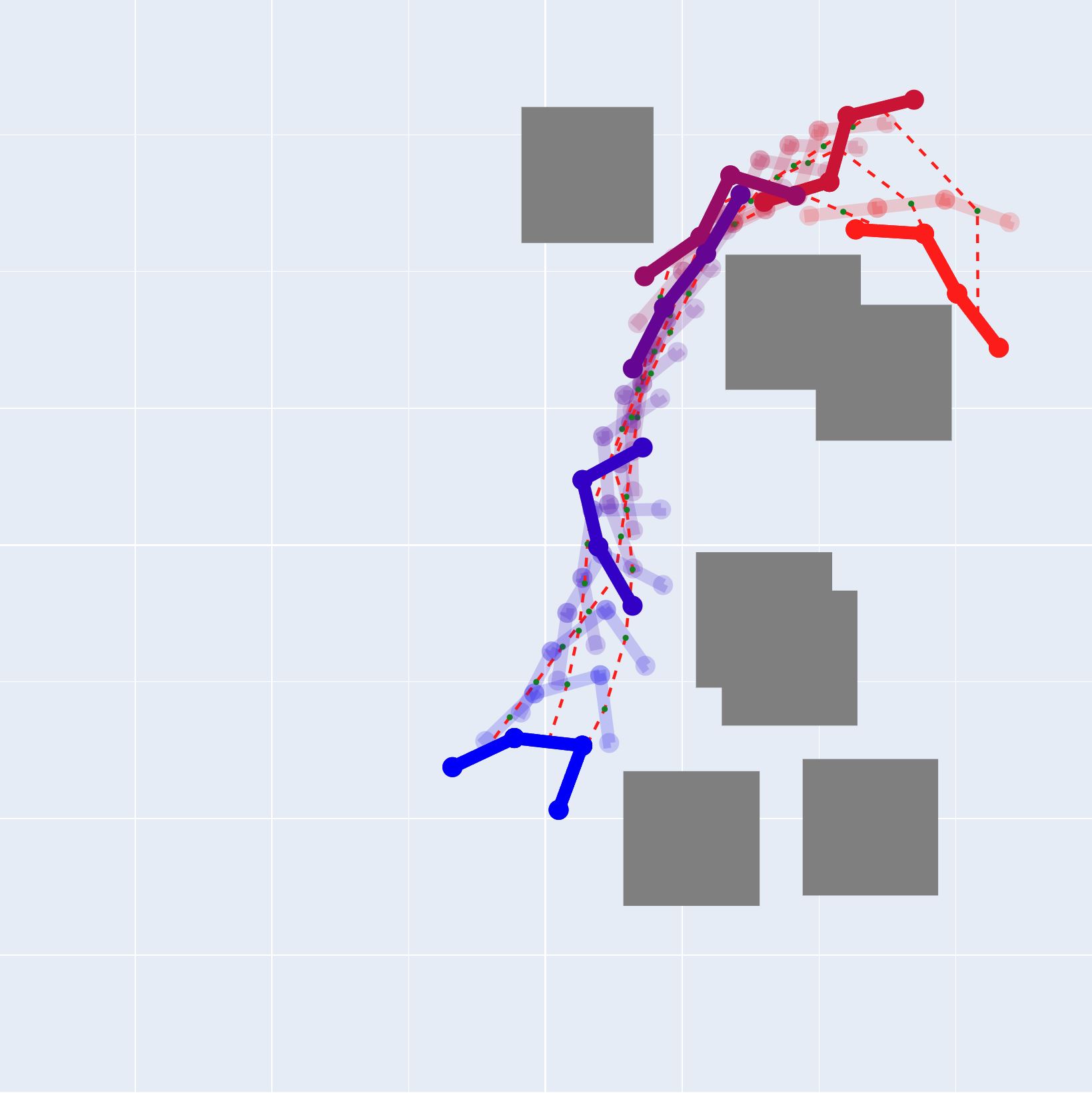}
}
\subfigure[3D point.]{
\includegraphics[width=1.2in]{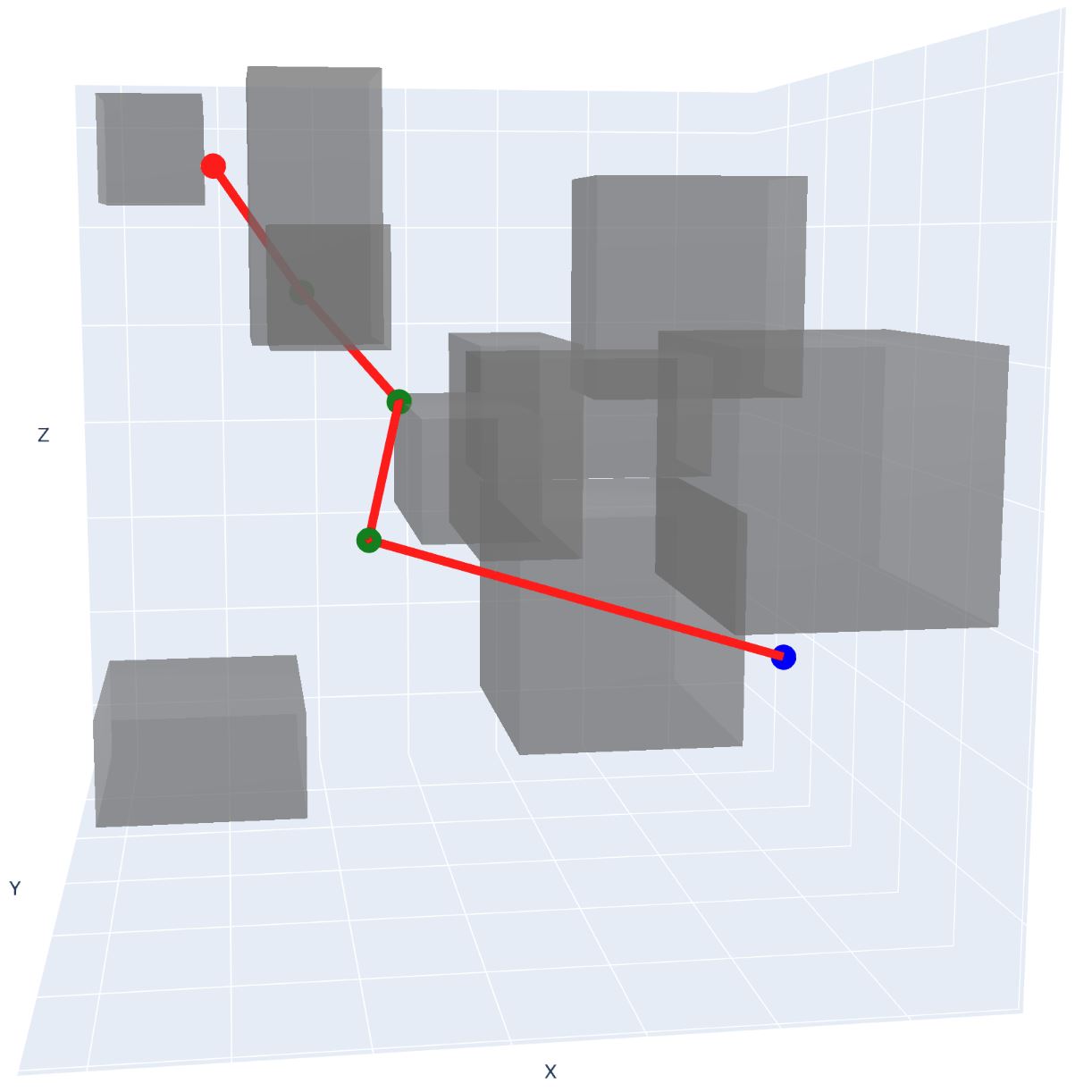}
}
\caption{More cases. Paths planned by MNP with different environment setup.}
\label{fig:multi strcuture appendix}
\end{figure*}

\begin{small}
    \begin{table*}[!ht]
\center
\begin{tabular}{|c|c|c|c|c|c|c|c|c|c|c|}\hline
\multicolumn{1}{|c|}{\multirow{3}{*}{Method}} & \multicolumn{10}{c|}{Environment \& Robot Model}\\ \cline{2-11}
  &\multicolumn{2}{|c|}{\multirow{1}{*}{2D Point}}& \multicolumn{2}{|c|}{\multirow{1}{*}{2D Rigid}}& \multicolumn{2}{|c|}{\multirow{1}{*}{2D 2-link}}& \multicolumn{2}{|c|}{\multirow{1}{*}{2D 3-link}}& \multicolumn{2}{|c|}{\multirow{1}{*}{3D Point}}\\ \cline{2-11}
& seen & unseen& seen& unseen& seen& unseen& seen& unseen& seen& unseen\\ \hline
MPNet(NR)    & 0.953&1.115 & 0.953&0.953 & 0.988&0.981 & 1.003&1.004 & 0.992&0.987 \\ \hline
MNP(Origin)  & 0.956&0.956 & 0.937&0.936 & 1.020&1.100 & 1.037&1.040 & 1.016&1.022 \\ \hline
MPNet(HR)    & 1.139&1.115 & 1.230&1.199 & 1.091&1.106 & 1.090&1.085 & 1.111&1.110 \\ \hline
MNP(RRT)     & \textbf{0.981}&\textbf{0.974} & \textbf{0.955}&\textbf{0.953} & 1.076&1.149 & 1.098&1.070 & 1.036&1.044 \\ \hline
IRRT*        & 1.002&1.002 & 0.987&0.988 & \textbf{1}&\textbf{1}         & 1.010&1.030 & 1.010&1.010 \\ \hline
BIT*      & 1&1         & 1&1         & \textbf{1}&\textbf{1}        & \textbf{1}&\textbf{1}           & \textbf{1}&\textbf{1}         \\ \hline
\end{tabular}
\caption{Mean path length comparison results. BIT* is the baseline. Note that only feasible paths are used for the length comparison. Since MPNet(NR) and MNP(Origin) do not have RRT as their replanner, they may fail in some challenging tasks which makes the length relatively short. We focus mainly on the comparison of the four remaining algorithms.}
\label{tab:length}
\end{table*}
\end{small}

\begin{figure*}[htbp]
\centering
\hspace{-0.4cm}
\subfigure[Path B step 1.]{
\includegraphics[width=1.1in]{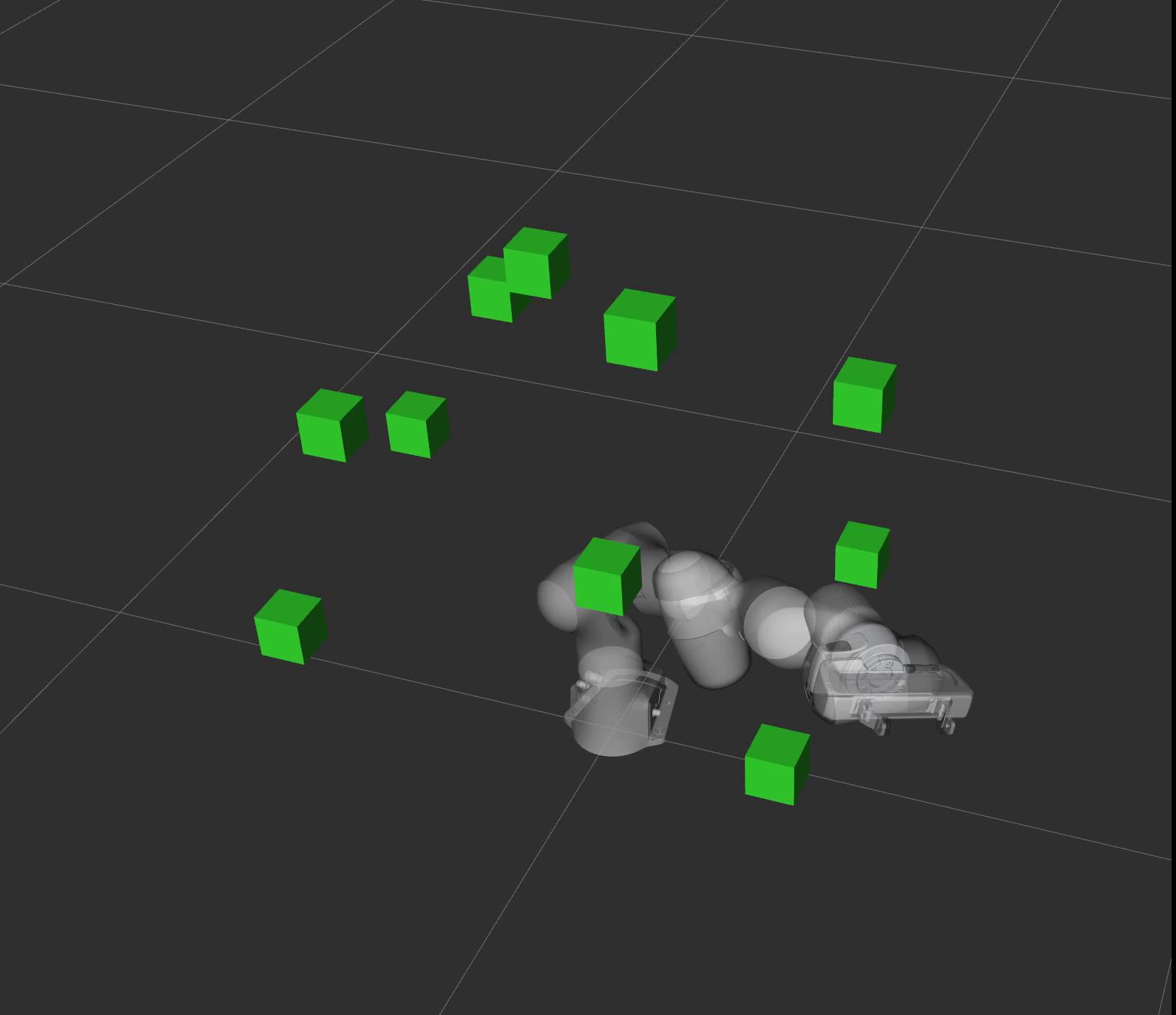}
}
\hspace{-0.4cm}
\subfigure[Path B step 2.]{
\centering
\includegraphics[width=1.1in]{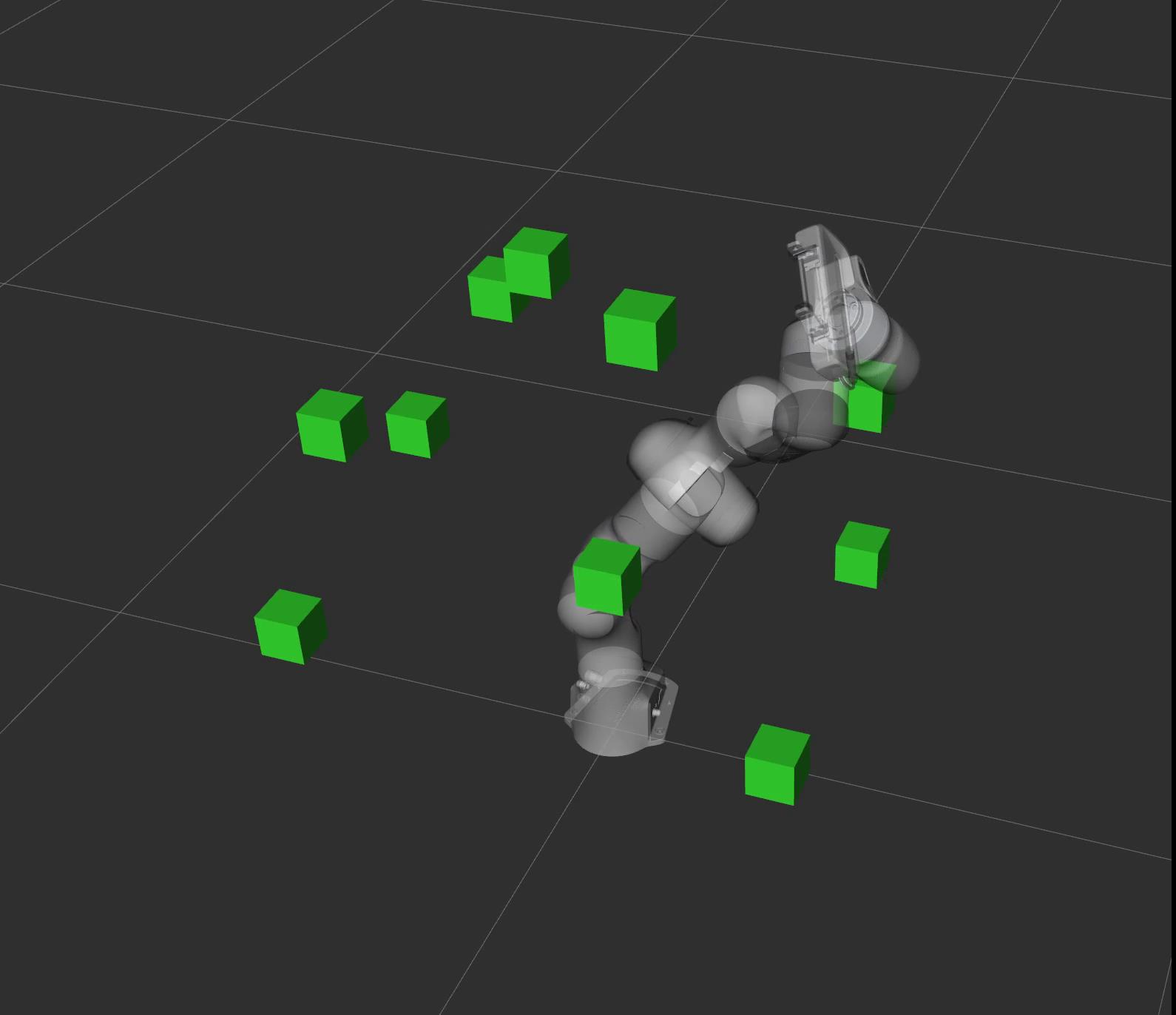}
}
\hspace{-0.4cm}
\subfigure[Path B step 3.]{
\centering
\includegraphics[width=1.1in]{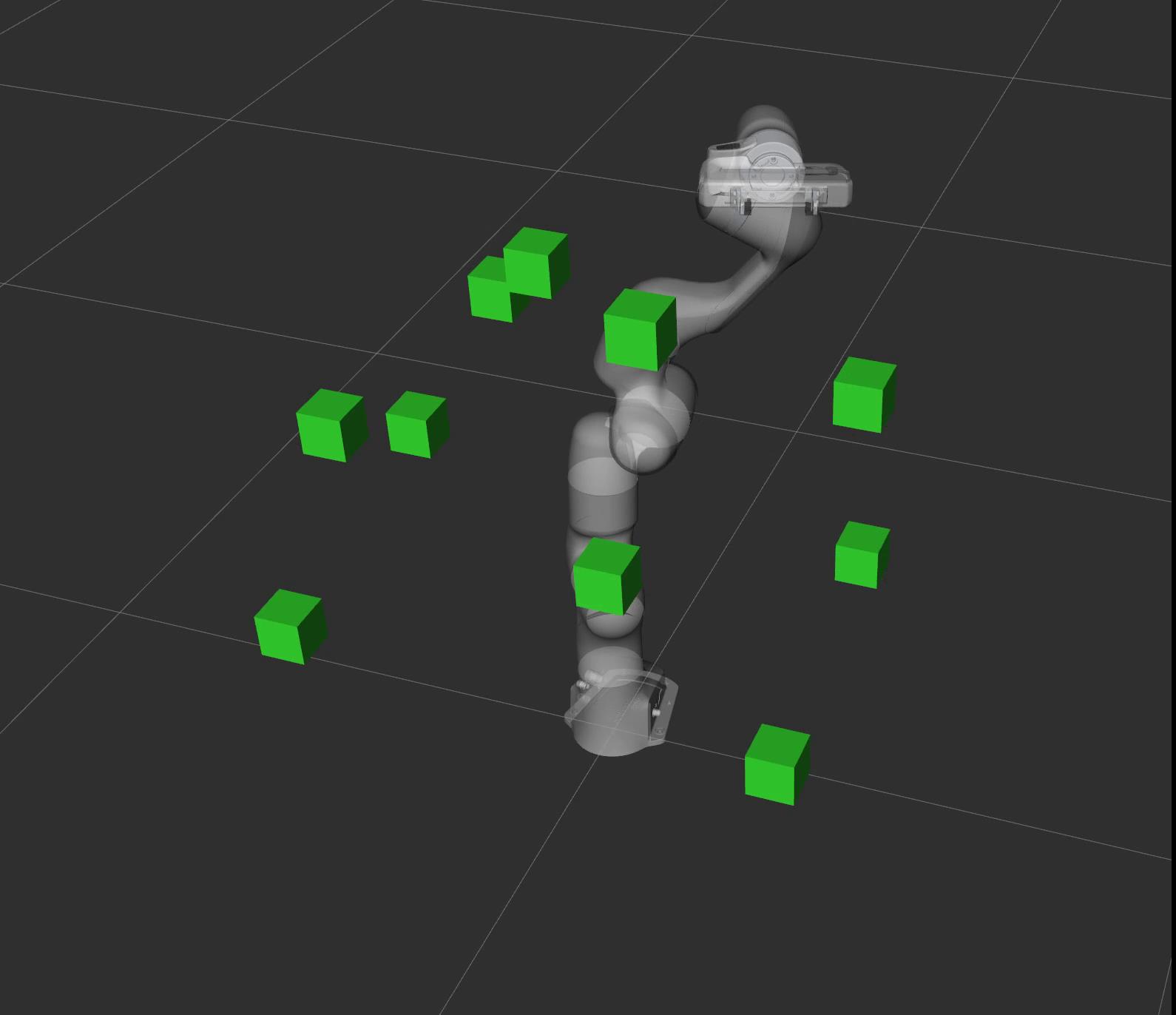}
}
\hspace{-0.4cm}
\subfigure[Path B step 4.]{
\centering
\includegraphics[width=1.1in]{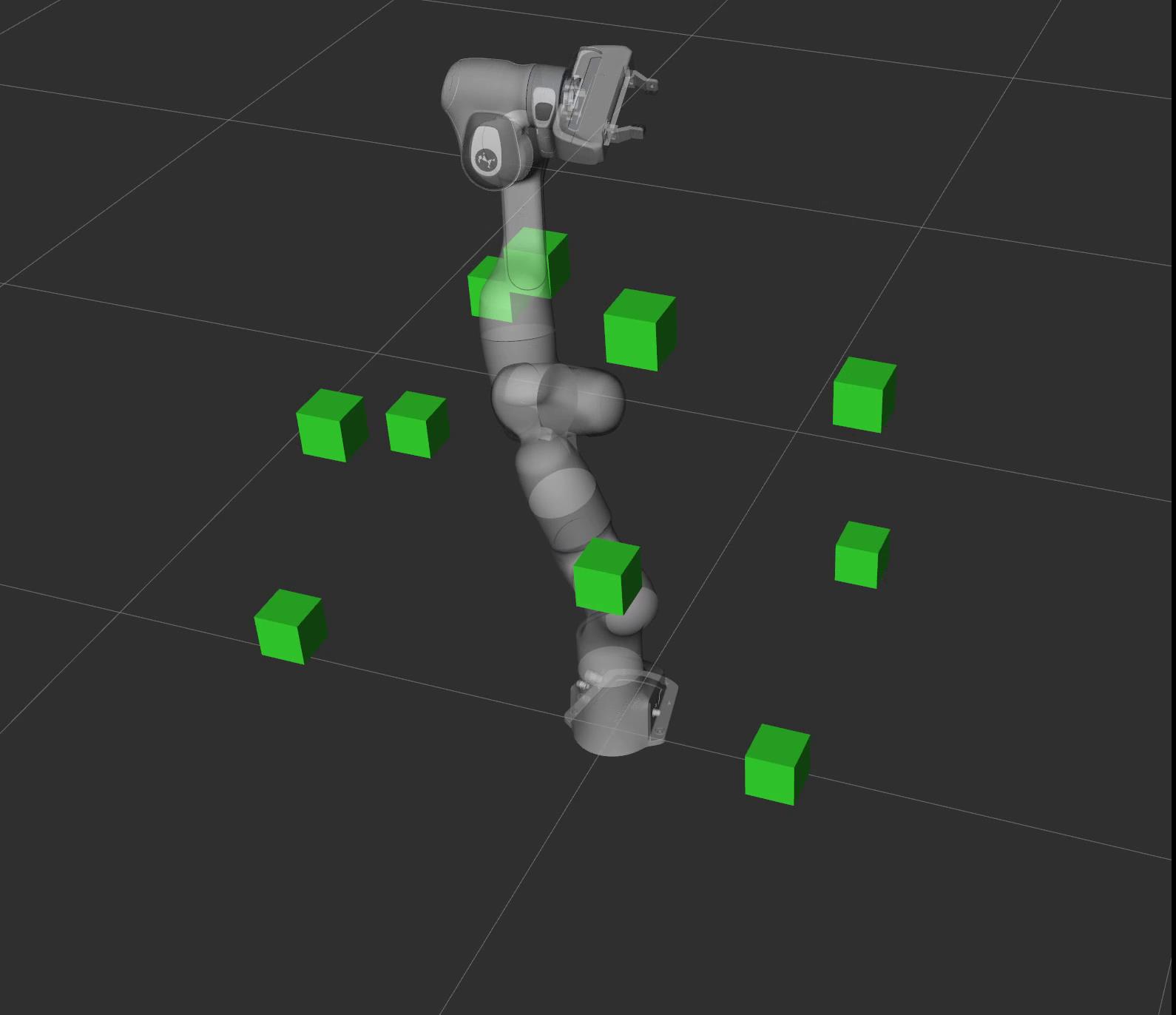}
}
\hspace{-0.4cm}
\subfigure[Path B step 5.]{
\centering
\includegraphics[width=1.1in]{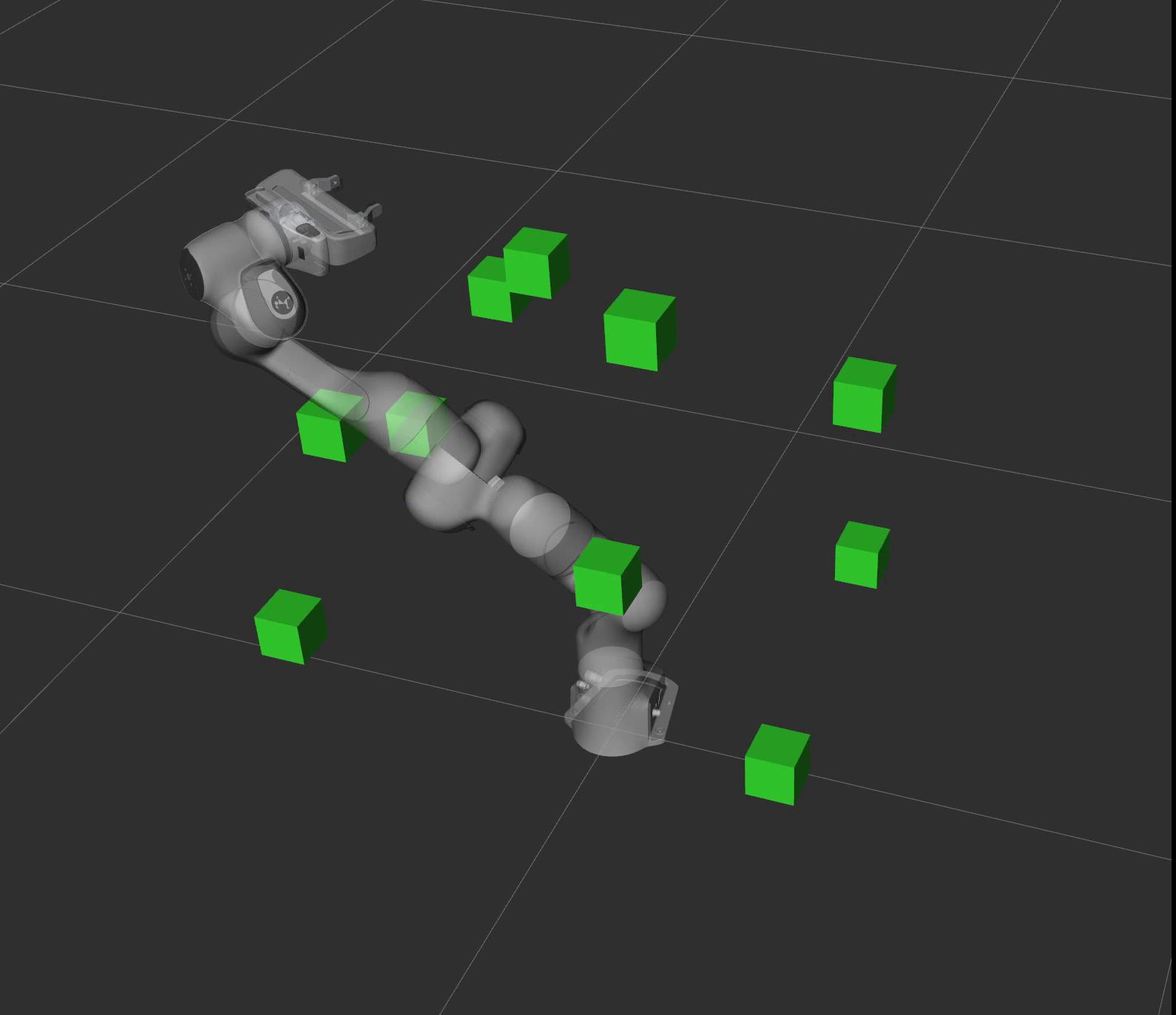}
}
\caption{Paths planed by MNP for the Panda Arm where green cubes are obstacles. 
}
\label{fig:arm appendix}
\end{figure*}

\par The supplemental materials to Sec~\ref{sec:comparison multiple} are presented here. Fig.~\ref{fig:multi strcuture appendix} shows more cases of the planned paths by MNP with different robot models and environments, and we can see that MNP is able to solve different motion planning problems and generalized well. Table~\ref{tab:length} shows the comparison results on the  lengths of generated paths. Note that BIT* is set as the baseline, and we stop IRRT* and BIT* when the length of the planned path is smaller than $110\%$ of the MNP(RRT)'s. Compared with MPNet(HR), MNP(RRT) finds shorter paths in 2D point, 2D Rigid and 3D Point environments, and it obtains paths of comparable length in 2D 2-link and 2D 3-link environments, with much less computation time. Sampling based methods such as BIT* and IRRT* also take much more computation time to get paths which have similar lengths compared with MNP(RRT)'s.

\bibliographystyle{IEEEtran}
\bibliography{reference}
\end{document}